\documentclass{article}

\usepackage{arxiv}

\usepackage[utf8]{inputenc} % allow utf-8 input
\usepackage[T1]{fontenc}    % use 8-bit T1 fonts
\usepackage{hyperref}       % hyperlinks
\usepackage{url}            % simple URL typesetting
\usepackage{booktabs}       % professional-quality tables
\usepackage{amsfonts}       % blackboard math symbols
\usepackage{nicefrac}       % compact symbols for 1/2, etc.
\usepackage{microtype}      % microtypography
\usepackage{lipsum}
\usepackage{graphicx}
\usepackage{tabularx}
\usepackage{amsmath}

\title{Deep Reinforcement-Learning-based Driving Policy for Autonomous Road Vehicles}

\author{
  Konstantinos Makantasis$^{1,3}$, Maria Kontorinaki$^{1,2}$, Ioannis Nikolos $^{1}$ \\
  $^1$ School of Production Engineering and Management, Technical University of Crete, Chania, Greece \\
  $^2$ Department of Statistics and Operations Research, Faculty of Science, University of Malta, Msida, Malta\\
  $^3$ Institute of Digital Games, University of Malta, Msida, Malta\\
  \texttt{konst.makantasis@gmail.com, mkontorinaki@dssl.tuc.gr, jnikolo@dpem.tuc.gr} 
}

\begin{document}
\maketitle

\begin{abstract}
In this work the problem of path planning for an autonomous vehicle that moves on a freeway is considered. The most common approaches that are used to address this problem are based on optimal control methods, which make assumptions about the model of the environment and the system dynamics. On the contrary, this work proposes the development of a driving policy based on reinforcement learning. In this way, the proposed driving policy makes minimal or no assumptions about the environment, since a priori knowledge about the system dynamics is not required. Driving scenarios where the road is occupied both by autonomous and manual driving vehicles are considered. To the best of our knowledge, this is one of the first approaches that propose a reinforcement learning driving policy for mixed driving environments. The derived reinforcement learning policy, firstly, is compared against an optimal policy derived via dynamic programming, and, secondly, its efficiency is evaluated under realistic scenarios generated by the established SUMO microscopic traffic flow simulator. Finally, some initial results regarding the effect of autonomous vehicles' behavior on the overall traffic flow are presented. 
\end{abstract}

% keywords can be removed

\section{Introduction}
In recent years, there has been a growing interest in self-driving vehicles. Building such autonomous systems has been an active area of research \cite{ziegler2014trajectory, cosgun2017towards} for its high potential in leading to road networks that are much more safer and efficient. Although vehicle automation has already led to great achievements in supporting the driver in various monotonous and challenging tasks, see for example \cite{reimer2010evaluation}, rising the level of automation to fully-automated driving is an extremely challenging problem. This is mainly due to the complexity of real-world environments, including avoiding obstacles, and human driving behavior aspects.

According to Donges \cite{donges1999conceptual}, autonomous driving tasks can be roughly classified into three categories; navigation, guidance, and stabilization. Navigation tasks are responsible for generating road-level routes. Tactical level guidance tasks are responsible for guiding autonomous vehicles along these routes in complex environments by generating tactical maneuver decisions. Finally, operational level stabilization tasks are responsible for translating tactical decisions into reference trajectories and then low-level controls that need to be tracked by the vehicle.

Several methodologies have been proposed for addressing the problem of generating efficient road-level routes. In the works of \cite{menelaou2017improved} and \cite{menelaou2017controlling} the authors propose a navigation algorithm based on a route reservation mechanism, in order to generate road-level routes, and at the same time avoid traffic congestion. In \cite{bast2010fast} a computationally efficient algorithm that can scale to very large transportation networks is presented. The authors of \cite{figliozzi2010vehicle} focus on "green" navigation by proposing a methodology to generate routes that minimize emissions. The work in \cite{baldacci2012recent} surveys exact algorithms for addressing the routing problem under capacity and travel time constraints. Finally, a comprehensive review regarding the generation of road-level routes in transportation networks can be found in \cite{bast2016route}. Despite the research interest, navigation examined by the autonomous driving perspective can be considered as a mature technology, since already exist commercial and free applications for road-level route generation.

At the same time, vehicles are man-made products for which automotive industry has decades-long experience in vehicle dynamics modeling, see for example \cite{gillespie1997vehicle} and \cite{rajamani2011vehicle}. Therefore, the operational level stabilization tasks, also know as the \textit{acting} part of autonomous driving, are well understood and modelled in control theory and robotics.   

Tactical level guidance, referred also as \textit{driving policy}, is crucial for enabling fully autonomous driving. On the contrary, however, to navigation and stabilization, tactical level guidance methodologies cannot be considered mature enough, in order to be applied to autonomous vehicles that move in unrestricted environments. A driving policy should be able to make decisions in real-time and in complex environments, in order to plan and update a vehicle path, which should be safe, collision-free, and user acceptable \cite{zhang2013dynamic}. The requirement for real-time operation in highly complex environments, as well as the safety constraints make current driving policies inadequate for fully autonomous driving. 

Based on the discussion so far, this work aims to contribute towards the development of a robust driving policy for autonomous vehicles that being capable of making decisions in real-time. The operation environment s restricted by considering vehicles that move on a highway. Highways consist a specific and very important type of transportation networks \cite{brilon2005reliability,yazici2014highway}. Due to their high-capacity they can serve millions of people everyday, while at the same time, allow users to travel with higher speed and less accelerations/decelerations compared to urban transportation networks. The driving policy development problem is formulated from an autonomous vehicle perspective (ego vehicle), and, thus, there is no need to make any assumptions regarding the kind of other vehicles (manual driving or autonomous) that occupy the road.

Finally, the proposed methodology approaches the problem of driving policy development by exploiting recent advances in \textit{Reinforcement Learning} (RL) combined with the responsibility sensitive safety model, proposed in \cite{shalev2017formal}. The developed RL-based driving policy aims to avoid accidents (departures from the road and crashes with other vehicles), move the vehicle with a desired speed, minimize accelerations/decelerations, and minimize lane changes. The latter two criteria are also related to the comfort of vehicle passengers \cite{ntousakis2016optimal}.

\subsection{Related Work}

The problem of path planning for autonomous vehicles can be seen as a trajectory generation problem corresponding to the creation of a quasi-continuous sequence of states that must be tracked by the vehicle over a specific time horizon. Trajectory generation has been widely studied in robotics \cite{goerzen2010survey}. Considering, however, road vehicles, path planning is a much more critical task, since passengers' safety must be guaranteed. 

Under certain assumptions, simplifications and conservative estimates, heuristic, hand-designed rules can be used for tactical decision making \cite{werling2008robust}. Such methods, however, are often tailored for specific non-complex environments and do not generalize robustly \cite{fletcher2008cornell}. Therefore, they are not able to cope with the complexity of real-world environments and the diversity of driving conditions, let alone human driving behavior aspects. To overcome the limitations of rule-based methods, approaches based on the careful design and exploitation of \textit{potential field} and \textit{optimal control} methods have also been proposed.

Potential field methods generate a field, or, in other words, an objective function the minimization of which corresponds to the objectives of an agent. These methods are based on the design of potential functions for obstacles, road structures, traffic regulation and the goals to be achieved. Then, the overall objective function is expressed as the weighted sum of the designed potential functions. The minimization is achieved via the generation of a vehicle trajectory moving towards the descent direction of the overall objective function \cite{wolf2008artificial, wang2015driving, schildbach2015scenario}. However, due to the fact that vehicle dynamics are not considered during decision making, the generated trajectory may turn out to be non-feasible to be tracked by the vehicle \cite{erlien2015shared}.  

This drawback can be alleviated by formulating the trajectory generation problem as an optimal control problem, which inherently takes into consideration system dynamics. Specifically, optimal control approaches allow for the \textit{concurrent} consideration of system dynamics and carefully designed potential fields \cite{zhang2016optimal}. In the work of  \cite{ntousakis2016optimal} an optimal control methodology for vehicles' trajectory planning in the context of cooperative merging on highways, is presented. The works of \cite{carvalho2014stochastic, gao2014tube} propose two optimal control based methodologies for trajectory planning, which incorporate constraints for obstacles, so as to keep the automated vehicle robustly far from them. In the same spirit, the works in \cite{makantasis2018motorway} and in \cite{gao2010predictive} design appropriate potential functions corresponding to the presence of obstacles, which, in turn, are incorporated in the objective function to generate a collision-free path. Optimal control approaches usually map the optimal control problem to a nonlinear programming (NLP) problem that can be solved using numerical NLP solvers, see for example \cite{gray2012semi, werling2012automatic, makantasis2018motorway}. Although, potential field and optimal control methods are quite popular due to the intuitive problem formulation \cite{rasekhipour2017potential}, there are still open issues regarding the decision making process.

First of all, mapping the optimal control problem to a NLP problem and solving it by employing numerical NLP solvers, produces a locally optimal solution for which the guarantees of the globally optimal solution may not hold, and, thus, the safety guarantees for the generated trajectory may be compromised \cite{papageorgiou2016feasible}. For this reason dynamic programming techniques have also been proposed for solving the optimal control problem. Although, dynamic programming techniques produce a globally optimal solution, due to the \textit{curse of dimensionality} \cite{bellman1952theory}, they are restricted to small scale problems. Moreover, another problem faced with potential field and optimal control approaches is the strong dependency to a relatively simple environment model, usually with hand-crafted observation spaces, transition dynamics and measurements mechanisms. These assumptions limit the generality of these methods to complex scenarios, since they are not able to cope with environment uncertainties and measurements errors. Finally, optimal control methods are not able to generalize, i.e., to associate a state of the environment with a decision without solving an optimal control problem. This means that every time a sequence of decisions needs to be made an optimal control problem needs to be solved, even if exactly the same problem has been solved in the past. This requirement significantly increases the computational cost of these methods.  

Due to its recent success, supervised deep learning has also been considered as an alternative approach for developing driving policies. In \cite{bojarski2017explaining} a convolutional neural network is trained in a supervised manner to output continuous steering actions. In the work of \cite{chen2017brain} a recurrent neural network is trained to output a steering angle, after a driving intention has been estimated. The works in \cite{chen2015deepdriving} and \cite{xu2017end} also exploit end-to-end trainable neural networks that output feasible driving actions and affordance indicators (such as distance between cars). The aforementioned approaches are based on end-to-end trainable neural network architectures that are able to output low-level controls directly from input images. Therefore, this kind of driving policies correspond to the outcome of a supervised learning algorithm, where deep neural networks were trained to imitate the behavior of human drivers. However, such methods, first, result to black-box driving policies, which are susceptible to influence of drifted inputs, and second, are restricted to the limitations of end-to-end learning \cite{glasmachers2017limits}. 

Very recently, RL methods have also been proposed as challenging alternative approaches towards the development of driving policies. RL-based approaches alleviate the strong dependency on hand-crafted simple environment models and dynamics, and, at the same time, can fully exploit the recent advances in deep supervised machine learning \cite{mnih2015human}. Along this line of research the work \cite{isele2017navigating} utilizes a deep Q-Network to make decisions for intersection crossing, while the work \cite{mukadam2017tactical} exploits a similar architecture to make decisions about lane changing in freeways. In \cite{paxton2017combining}, the authors propose a hierarchical RL-based approach for deriving a low-level driving policy capable of guiding a vehicle from an origin point to a destination point. In \cite{shalev2016safe} a policy gradient RL approach is used to develop a driving policy for cooperative double merging scenarios. This approach combines a RL policy with a non-learnable mechanism to balance between efficiency and safety. Finally, the work in \cite{liu2018elements} presents some elements of efficient deep RL (empirically validated) for decreasing the learning time and increasing the efficiency of RL-based driving policies.   

Despite the fact that only very recently reinforcement learning was employed for developing driving policies, experimental results appear very promising. The main drawback, however, of these approaches regards safety guarantees. Due to the fact that the probability of an accident is very small, learning based approaches, as shown in \cite{shalev2017formal}, cannot assure collision-free trajectories.

\subsection{Proposed Work}
This work proposes a RL-based approach towards the development of a driving policy for autonomous road vehicles. The proposed RL-based method has several advantages over potential field and optimal control methods. First of all, RL-based approaches are \textit{model-free}. They make the assumption that there is a state-transition model that describes the system dynamics, which remains fixed. However, the exact form of this model is not required to be a priori known (typically such a model is considered unknown), but it is being inferred during training. Second, a driving policy based on RL is able to \textit{generalize}. After training, a RL-based policy has inferred a mapping for associating a given state of the environment with a decision. In contrast to potential field and optimal control methods, whenever a decision needs to be made no problem needs to be solved; decision making can be done by simply evaluating the policy function. Third, since a RL-based driving policy has been estimated, it can be shared across multiple autonomous vehicles, which in turn can make decisions through the policy function evaluations. On the contrary, driving policy sharing is not possible when potential field and optimal control methods are used, since each vehicle needs to solve a decision making problem for its own sake. Finally, since no learning-based driving policy can guarantee absolute safety, our work is motivated by the formal responsibility sensitive safety model, proposed in \cite{shalev2017formal}, in order to derive and utilize ad-hoc rules that guarantee responsibility-wise safety. That is, the ad-hoc rules guarantee that the autonomous vehicles will not be responsible for any occurred accident. To the best of our knowledge, this work is one of the first attempts that try to derive a RL driving policy, combined with ad-hoc safety rules, targeting unrestricted highway environments, which are occupied by both autonomous and manual driving vehicles.   

Furthermore, the proposed RL-based driving policy is compared against an optimal policy derived using dynamic programming, in terms of safety metrics, such as the number of collisions, and efficiency metrics, such as the average time the autonomous vehicle moves with the desired speed. Although, dynamic programming techniques, due to the curse of dimensionality \cite{bellman1954theory}, are restricted to small-scale problems, and are not suitable for real-time applications, they produce globally optimal solutions to an optimal control problem, i.e., optimal driving policies. Thus, the comparison of the proposed methodology against optimal driving policies, first, will result to an objective evaluation for the RL-based driving policy, and, second, can provide insights to the driving policy development problem.    

The developed RL-based driving policy is also compared against manual driving using SUMO simulator. Through this comparison, the generalization ability and stability of the proposed RL-based driving policy to ensure reliability is evaluated; any learning system must generalize well to out-of-sample data, and be stable, i.e., small perturbations in the input should slightly affect the output. Specifically, the RL-based driving policy is applied to randomly generated driving scenarios (previously unseen driving conditions), with and without drivers' imperfection and measurements errors. Drivers' imperfection and measurements errors can be seen as disturbances, and can be incorporated into driving scenarios using appropriate settings in SUMO simulator. 

Finally, preliminary results regarding the effect of autonomous vehicles on the overall traffic flow are provided. The RL-based driving policy, seen by an autonomous vehicle perspective, is a selfish policy. That is, each autonomous vehicle that follows the RL policy tries to achieve its own goals disregarding the rest of the vehicles. Such a behavior might have a negative effect on the overall traffic flow. 

The rest of the paper is organized as follows: Section \ref{sec:problem_formulation} describes the problem and the underlying assumptions, Section \ref{sec:RL-based_driving_policy} gives a brief description of the RL framework, Section \ref{sec:RL-based_driving_policy} presents in detail the development of the RL-based driving policy and Section \ref{sec:rules} the derivation of ad-hoc rules towards the design of a collision-free trajectory. Section \ref{sec:4} presents the experimental setup and the experimental results, and Section \ref{sec:5} concludes this work.

\section{Problem Description and Assumptions}
\label{sec:problem_formulation}
The problem of path planning for an autonomous vehicle that moves on freeway, which is also occupied by manual driving vehicles is considered. Without loss of generality, it is assumed that the freeway consists of three lanes. The path planning algorithm, or in other, words the driving policy, should generate a collision-free trajectory for the autonomous vehicle to follow. Moreover, the generated trajectory should permit the autonomous vehicle to move forward with a desired speed, and, at the same time, minimize its longitudinal and lateral accelerations/decelerations. The aforementioned three criteria are the objectives of the driving policy, and therefore, the goal that the RL algorithm should achieve.

For the generation of an optimal trajectory using dynamic programming, the manual driving vehicles is required to move with a constant speed following the kinematics equations. The generation of the optimal trajectory, via dynamic programming, corresponds to the solution of a finite horizon optimal control problem. The aforementioned requirement assures that the dynamics of the system will be a priori and fully known, and no disturbances will be present in the system in order for the dynamic programming technique to produce the trajectory. However, for training the RL policy the aforementioned system dynamics are not given to the algorithm, and, thus, are considered unknown.

Regarding the SUMO simulator, the manual driving vehicles move on the freeway using the Krauss car following model \cite{kanagaraj2013evaluation}. It is assumed that letting the manual driving vehicles to move using the Krauss car following model will produce realistic driving behaviors. Moreover, manual driving vehicles should move forward with a desired speed. In order to generate realistic and customary traffic conditions, we assume that at least two categories of manual driving vehicles should be present at the freeway; manual driving vehicles that want to move faster than the autonomous vehicle, and manual driving vehicles that want to move slower. At this point, it should be stressed that, although the manual vehicles are moving using the Krauss model, this model is not given to the RL training algorithm, and, thus, from a RL point of view it is considered unknown.

During the trajectory generation this work does not assume any communication between the autonomous vehicle and other vehicles. Instead, the information available for the trajectory generation is obtained solely by sensors, such as cameras, LiDAR and proximity sensors, installed on the autonomous vehicle. This work also assumes the availability of a fusion module of the on-board sensors' information, with the appropriate redundancy and cross-checking, to assure the usefulness and accuracy of the provided information. Using such sensors, the autonomous vehicle can estimate the position and the velocity of its surrounding vehicles. Therefore, the state representation of the autonomous vehicle and its surrounding environment, includes information that is associated solely with the position and the velocity of the vehicles present in the sensing area of the autonomous vehicle. 

Furthermore, it is assumed that the freeway does not contain any turns. However, the generated vehicle trajectory essentially reflects the vehicle longitudinal position, speed, and its traveling lane. The derived trajectory needs to be tracked by the underlying vehicle control loops, based on high-definition maps. Therefore, for the trajectory specification, possible curvatures may be aligned to form an equivalent straight section \cite{makantasis2018motorway}. 

Finally, the trajectory of the autonomous vehicle can be fully described by a sequence of goals that the vehicle should achieve. Each one of the goals should be achieved within a specific time interval, and represents vehicle's desires, such as change lane, brake with a given deceleration, etc. These goals define the trajectory to be followed by the autonomous vehicle in a higher level, and cannot be directly used by the vehicle control loops. Instead, it is assume that the mechanism which translates these goals to low level controls and implements them is given.

Based on the aforementioned problem description and underlying assumptions, the main objective of this work is to develop a driving policy. The driving policy will exploit the information coming from a set of sensors installed on the autonomous vehicle, in order to set a goal for the vehicle to achieve, via a high-level action, during a specific time interval. In other words, the objective is to derive a function that will map the information about the autonomous vehicle, as well as, its surrounding environment to a specific goal and the corresponding high-level action for achieving it.

\section{RL and Prioritized Experience Replay}
\label{sec:RL-based_driving_policy}
In this work the development of a driving policy is being tackled as a RL problem, where the state-action value function $Q$ is approximated by a Double Deep Q-Network (DDQN) \cite{van2016deep} using prioritized experience replay \cite{schaul2015prioritized}. Therefore, for the sake of completeness, in this section the RL framework and the algorithm of prioritized experience replay are briefly presented.

\subsection{Reinforcement Learning}
In the RL framework, an agent interacts with the environment in a sequence of actions (selected by following a specific policy), observations, and rewards. In particular, at each time step $t$, the agent (in our case the autonomous vehicle) observes the state of the environment $s_t \in \mathcal S$ and, based on a specific policy, it selects an action $a_t \in \mathcal A$, where $\mathcal S$ is the state space and $\mathcal A=\{1,\cdots, K\}$ is the set of available actions. Then, the agent observes the new state of the environment, $s_{t+1}$, which is the consequence of applying the action $a_t$ at state $s_t$, and a scalar reward signal $r_t$, which is a quality measure of how good is to select action $a_t$ at state $s_t$.

The goal of the agent is to interact with the environment by selecting actions in a way that maximizes the cumulative future rewards, also known as \textit{future return}. Future rewards are discounted by a factor $0\leq \gamma < 1$ per time step, and the future return at time $t$ is defined as
\begin{equation}
\label{eq:1}
R_t = \sum_{t'=t}^{T}\gamma^{t'-t}r_{t'},
\end{equation} 
where parameter $T$ denotes how many time steps ahead $t$ are taken into consideration for calculating $R_t$. The non-negative discount factor $0 \leq \gamma < 1$ determines the importance of future rewards. In other words, it weighs future rewards, by giving higher weight to rewards received near than rewards received further in the future.

The interaction of the agent with the environment can be explicitly defined by a policy function $\pi:\mathcal S \rightarrow \mathcal A$ that maps states to actions. The maximum expected future reward achievable by following any policy after observing a state $s$ and selecting an action $a$ is represented by the optimal \textit{action-value} function $Q^*(s,a)$, which is defined as
\begin{equation}
Q^*(s,a) = \max_{\pi}\mathbb E [R_t | s_t=s, a_t=a, \pi].
\end{equation} 
The optimal action-value function obeys a very important identity known as \textit{Bellman equation}. That is, if the optimal action-value function $Q^*(s_{t+1},a_{t+1})$ of the state $s_{t+1}$ at the next time step was known for all possible actions $a_{t+1}$, then the policy maximizing the future reward is to select the action maximizing the expected value of $r + \gamma Q^*(s_{t+1},a_{t+1})$, and, thus, the following
\begin{equation}
\label{eq:bellman_update}
Q^*(s,a) = \mathbb E_{s_{t+1}} [r_t+ \gamma \max_{a_{t+1}} Q^*(s_{t+1},a_{t+1}) | s_t=s, a_t=a]
\end{equation}
holds for the optimal action-value function when state $s_t$ is observed and action $a_t$ is selected. The expectation in relation (\ref{eq:bellman_update}) is with respect to all possible states at the next time step. 

The relation in (\ref{eq:bellman_update}) implies that the problem of estimating the optimal policy is equivalent to the estimation of $Q^*(s,a)$ for every pair $(s,a) \in \mathcal S \times \mathcal A$. Although, $Q^*(s,a)$ can be efficiently estimated when small scale problems need to be addressed \cite{watkins1992q}, for large state spaces estimating $Q^*(s,a)$ for every possible $(s,a)$ pair is practically implausible. For such kind of problems, the optimal action-value function is approximated, $\tilde{Q}^*(s,a;\theta) \approx Q^*(s,a)$, using a learning machine, such as linear regression of neural networks \cite{mnih2015human}, parameterized by $\theta$. Parameters $\theta$ are estimated by following an iterative procedure for minimizing a sequence of loss functions
\begin{equation}
\label{eq:loss_functions}
L_i(\theta_i) = \mathbb E_{s,a} [(\tilde{Q}(s,a;\theta_{i-1}) - \tilde{Q}(s,a;\theta_i))^2],
\end{equation}
where $i$ stands for the iteration index. The $(s,a,r,s')$ tuples used in relations (\ref{eq:bellman_update}) and (\ref{eq:loss_functions}) are generated by following an $\epsilon$-greedy policy that selects at a given state a greedy action with probability $1-\epsilon$ and a random action with probability $\epsilon$.

The aforementioned procedure for estimating $\theta$ looks like a regression problem in the supervised learning paradigm. However, there are two significant differences. First, the learning machine sets itself and follows the targets $\tilde{Q}(s,a;\theta_{i-1})$, which can lead to instabilities and divergence, and, second, the generated $(s,a,r,s')$ tuples are not independently generated; a property that is required by many learning machines. 

To overcome the first problem, two identical learning machines are used; one for setting the targets and one for following them. The machine that sets the targets is freezed in time, i.e., its parameters are fixed for several iterations. After a predefined number of iterations has passed, the parameters of the machine that sets the targets are updated by coping the parameters from the machine that follows the targets. If we denote as $\hat{\theta}$ the parameters of the machine that sets the targets, then the loss function $L_i(\theta_i)$ in relation (\ref{eq:loss_functions}) is given by
\begin{equation}
L_i(\theta_i) = \mathbb E_{s,a} [(\tilde{Q}(s,a;\hat{\theta}) - \tilde{Q}(s,a;\theta_i))^2].
\end{equation}

\subsection{Prioritized Experience Replay Algorithm}
To overcome the latter problem, a \textit{Prioritized Experience Replay} (PER) algorithm is employed to break the correlations between the generated $(s,a,r,s')$ tuples. The generated tuples are stored into a memory, and for minimizing (\ref{eq:loss_functions}) a training set $\mathcal D =\{(a,s,r,s')_j\}_{j=1}^n$ is drawn from the memory according to a distribution that prefers tuples that do not fit well to the current estimate of the action-value function.  

For estimating the sampling distribution, initially, the difference
\begin{equation}
d(s,a,r,s') = |\tilde{Q}(s,a;\hat{\theta}) - \tilde{Q}(s,a;\theta_i)| 
\end{equation} 
is computed for each tuple in memory and is updated after each iteration $i$. Then, the difference is converted to priority
\begin{equation}
p = (d - \epsilon)^a,
\end{equation}
with $\epsilon > 0$ to ensure that no tuple has zero probability of being drawn, and $0 \leq a < 1$ (when $a=0$ the uniform distribution over tuples is used). Finally, the priorities are translated into probabilities. In particular, a tuple $k$ has a probability 
\begin{equation}
\label{eq:probability}
P_k = \frac{p_k}{\sum_{j=1}^N p_j}
\end{equation}
of being drawn during the experience replay. Variable $N$ in (\ref{eq:probability}) stands for the cardinality of memory.

\begin{figure}[t]
	\begin{minipage}{1.0\linewidth}
		\centering
		\centerline{\includegraphics[width=0.98\linewidth]{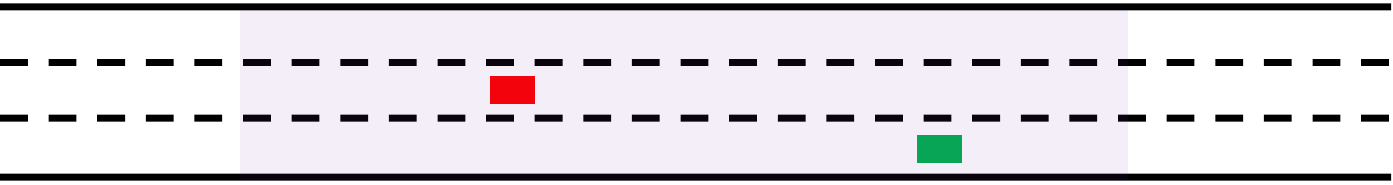}}
		\centering (a)
		\vspace{0.1in}
	\end{minipage} 
	\begin{minipage}{1.0\linewidth}
		\centering
		\centerline{\includegraphics[width=0.98\linewidth]{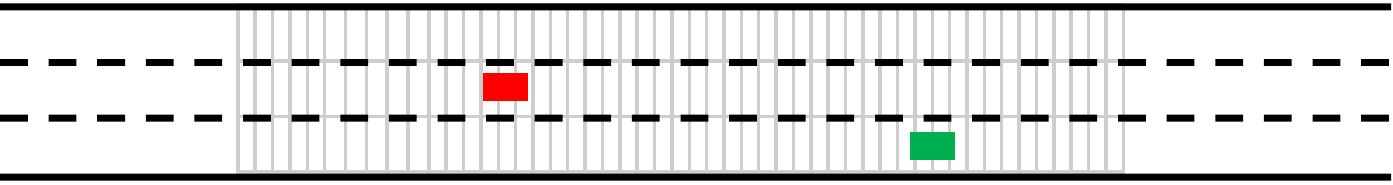}}
		\centering (b)
	\end{minipage} 
	\caption{State representation. The autonomous vehicle is represented by the red rectangle, while the green rectangle represents another vehicle present on the road. (a) The purple shaded area corresponds to the sensed surrounding environment of the autonomous vehicle. (b) Discretization of sensed environment.}
	\label{fig:state}
\end{figure}

\section{Driving Policy}
\label{sec:driving_policy}
Having described the RL framework and the the prioritized experience replay algorithm, in this section, the RL-based approach utilized in this work is presented, along with the state and action representation, and the design of the scalar reward signal. Finally, the architecture of the employed neural network, and details about the implementation, as well as, the mechanism for generating $(s,a,r,s')$ tuples for training the neural network are described.    

\subsection{State Representation}
Autonomous vehicles are equipped with multiple sensors that enable them to capture heterogeneous and multimodal information about their surrounding environment. This allows for a wide variety of state representations. The selection, however, of the representation significantly affects the ability of an agent to learn. In this work, a state representation that, on the one hand, can be constructed using current sensing technologies, and, on the other, it allows the agent to efficiently learn is utilized. 

Specifically, this work considers autonomous vehicles that move on a freeway with three lanes. It is assumed that the vehicle can sense the surrounding environment that spans 60 meters behind it and 100 meters ahead of it, as well as, its two adjacent lanes. This means that the autonomous vehicle can estimate the relative positions and velocities of other vehicles that are present in the aforementioned area. Note that with current LiDAR and camera sensing technologies such an assumption can be considered valid. A schematic representation of the sensed surrounding environment of a vehicle is presented in Figure \ref{fig:state}(a).

In order to translate the information that can be sensed by the autonomous vehicle into a state vector, the sensed area is discretized into tiles of one meter length, see Fig. \ref{fig:state}(b). In order for this discretization to be useful, the accuracy of the vehicle sensors must be in the order of centimeters, something that is feasible with current sensing technologies \cite{akai2017autonomous, wolcott2017robust, de2018fusion}. The value of the longitudinal velocity of the autonomous vehicle is assigned to the tiles beneath of it. To tiles occupied by other vehicles the value of their longitudinal velocity is assigned. The velocity of the other vehicles is estimated by using their positions in two subsequent time instances. The value of zero is given to all non occupied tiles that belong to the road and, finally, the value minus one to tiles outside of the road (the autonomous vehicle can sense an area outside of the road in case it occupies the left-most or the right-most lane).

Using the representation above, the sensed environment is transformed into a matrix with three rows and one hundred sixty columns. Moreover, this matrix contains information about the absolute velocities of vehicles, as well as, relative positions of other vehicles with respect to the autonomous vehicle. Finally, the vectorized version of this matrix, that is a vector with 480 elements, is used to represent the state of the environment at each specific time step.

The proposed state representation can be easily obtained by sensors installed on an autonomous vehicle. Despite the fact that this representation is relatively simple, as it can be seen in Section \ref{sec:4}, it contains adequate information for obtaining robust driving policies. More realistic and informative state representations can be constructed. For example current pattern and object recognition methods can be utilized to classify vehicles and, thus, incorporate into the state representation information regarding the type of surrounding vehicles and their size. In addition, if we assume that vehicles are equipped with communication enabling technologies, then vehicle-to-vehicle communication can be used to enhance the state representation with information regarding vehicles' longitudinal and lateral  accelerations, while vehicle-to-infrastructure communication can provide information regarding the state of the network. Although, more accurate state representations can be constructed, using a simple state representation, like the proposed one, permits to gain insights with respect to the behavior of the derived policy. Moreover, we deliberately do not assume any communication between the vehicles, to make the training of the RL policy much harder, and, at the same time, be able to evaluate its behavior under minimal assumptions. Finally, this work is based on the argument that RL based techniques can be proved very valuable towards the developments of driving policies, even in mixed driving scenarios, and thus, it can be seen as a preliminary proof-of-concept.

\subsection{Action Representation}
\label{ssec:actions}
Seven available actions are defined; i) change lane to the left, ii) change lane to the right, iii) accelerate with a constant acceleration of $1m/s^2$ or $2m/s^2$, iv) decelerate with a constant deceleration of $-1m/s^2$ or $-2m/s^2$, and v) move with the current speed at the current lane, see Table \ref{table:actions}. For the acceleration and deceleration actions feasible acceleration and deceleration values are used to ensure that the autonomous vehicle will be able to implement them. Moreover, the autonomous vehicle is making decisions by selecting one action every one second, which implies that the first two actions are also feasible, that is, a moving car is able to change lane in a time interval of one second. 

Using the aforementioned action representation, each action can be seen as a goal or desire of the autonomous vehicle that should be achieved during one second. Practically, the first six actions represent goals that are associated with the avoidance of obstacles. The third to sixth actions represent also goals that are related to the fact that the autonomous vehicle should move forward with a desired speed. Finally, the seventh action implies that the vehicle is moving with the desired speed and there are no obstacles to avoid.

Note that the goal of this work is to develop a driving policy by approximating through RL the action-values $Q(s,a)$ for every possible $(s,a)\in \mathcal S \times \mathcal A$ pair. Therefore, adopting an action space with small cardinality can significantly simplify the problem leading to faster training. Moreover, the authors of \cite{liu2018elements} argue that low-level control tasks can be less effective and/or robust for high-level driving policies. For these reasons an action space like the one presented above is used, instead of lower level commands such as longitudinal and lateral accelerations.

Finally, by using an action set of goals, the RL-based driving policy makes high-level decisions for leading the autonomous vehicle to a desired state. The implementation of these goals can efficiently take place by exploiting a separate non-learnable module, such as dynamic programming. This low-level module will produce state trajectories by translating each specific desire to lower level commands, such as longitudinal and lateral accelerations. These state trajectories may then be used as a reference by the vehicle throttle and brake controllers, which are designed on the basis of vehicle dynamics, to produce the actual vehicle movement on the road. As mentioned in Section \ref{sec:problem_formulation}, the development of such a module is beyond the scope of this work, and, thus, it is assumed that is given.

\begin{table}[t]
	\centering
	\caption{The available actions of the autonomous vehicle.}
	\newcolumntype{L}[1]{>{\hsize=#1\hsize\raggedright\arraybackslash}X}%
	\newcolumntype{C}[1]{>{\hsize=#1\hsize\centering\arraybackslash}X}%
	\label{table:actions}
	
	\begin{tabularx}{0.53\linewidth}{C{6.5}L{17.0}}
		\hline \hline 
		
		Action \#1: & Change lane to the left \\ \hline
		Action \#2: & Change lane to the right \\ \hline
		Action \#3: & Constant acceleration of $1m/s^2$ \\ \hline
		Action \#4: & Constant acceleration of $2m/s^2$ \\ \hline
		Action \#5: & Constant deceleration of $1m/s^2$ \\ \hline
		Action \#6: & Constant deceleration of $2m/s^2$ \\ \hline
		Action \#7: & Move on current lane with current speed \\ \hline
		\hline 
	\end{tabularx}
\end{table}

\subsection{Reward Signal Design}
The reward signal is a measure of the quality of a selected action at a specific state, and is the only mean through which a policy can be evaluated. So, designing appropriate rewards signals is the most important tool for shaping the driving behavior of an autonomous vehicle.

For driving scenarios, the autonomous vehicle should be able to avoid collisions, move with a specific desired speed and avoid unnecessary lane changes and accelerations. Therefore, the reward signal should reflect all these objectives by employing one penalty function for collision avoidance, one that penalizes deviations from the desired speed and two penalty functions for unnecessary lane changes and accelerations/decelerations.

The penalty function for collision avoidance should feature high values at the gross obstacle space, so that the autonomous vehicle is repulsed, and potentially unsafe decisions are suppressed; and low (or virtually vanishing) values outside that space. To this end, the exponential penalty function
\begin{equation}
f(\delta_i) = \begin{cases}
e^{-(\delta_i - \delta_0)} &\text{ if } l_e = l_i \\
0 &\text{ otherwise}
\end{cases}
\label{eq:distance_penalty}
\end{equation}
is adopted. In (\ref{eq:distance_penalty}) $\delta_i$ is the longitudinal distance between the automated vehicle and the $i$-th obstacle (the $i$-th vehicle in its surrounding environment), $\delta_0$ stands for the minimum safe distance, and, $l_e$ and $l_i$ denote the lanes occupied by the autonomous vehicle and the $i$-th obstacle, respectively. Note that this function is activated only when the automated vehicle and an obstacle are at the same lane. Finally, if the value of (\ref{eq:distance_penalty}) becomes greater or equal to one, then the driving situation is considered very dangerous and it is treated as a collision. 

The vehicle mission is to advance with a longitudinal speed close to a desired one. Thus, the quadratic term 
\begin{equation}
\label{eq:speed_penalty}	
h(v) = (v - v_d)^2
\end{equation}
that penalizes the deviation between the vehicle speed and its desired speed, is incorporated in the reward. In (\ref{eq:speed_penalty}) the variable $v$ stands for the longitudinal speed of the autonomous vehicle, while the constant $v_d$ represents its desired longitudinal speed. 

Two terms are also introduced; one for penalizing accelerations/decelerations, and one for penalizing unnecessary lane changes. For penalizing accelerations the term
\begin{equation}
a(v_t, v_{t-1}) = (v_t - v_{t-1})^2
\end{equation}
is used, while for penalizing lane changes the term
\begin{equation}
g(l_t, l_{t-1}) = \mathbb I(l_{t}\neq l_{t-1}).
\end{equation}
is used. Variables $v_t$ and $l_t$ correspond to the speed and lane of the autonomous vehicle at time step $t$, while $\mathbb I(\cdot$) is the indicator function.

The total reward at time step $t$ is the negative weighted sum of the aforementioned penalty terms, that is
\begin{equation}
r_t = -w_1 \sum_{i=1}^{O_t}f_t(\delta_i) - w_2 h_t(v_t) - w_3\sum_{i=1}^{O_t} \mathbb I(f_t(\delta_i) \geq 1- w_4 a(v_t,v_{t-1}) - w_5 g(l_t, l_{t-1})
\label{eq:reward}
\end{equation}
In (\ref{eq:reward}) the third term penalizes collisions and variable $O_t$ corresponds to the total number of obstacles that can be sensed by the autonomous vehicle at time step $t$. The selection of weights defines the importance of each penalty function to the overall reward. In this work the weights were set, using a trial and error procedure, as follows: $w_1=1$, $w_2=0.5$, $w_3=20$, $w_4=0.01$, $w_5=0.01$. The largest weighting factors are associated with the terms that penalize collisions and model obstacle avoidance, since the derived policy should generate collision free trajectories. The weighting term associated with the desired speed of the vehicle defines how aggressive and/or how conservative will be the derived driving policy. Using a small value for this weight will result to a conservative policy that will advance the vehicle with very low speed or, even worse, keep the vehicle immobilized by setting its speed equal to zero. Finally, the values of the weighting factors associated with lane changes accelerations/decelerations are small in order to enable the vehicle to make maneuvers, such as overtaking other vehicles.

\subsection{Neural Network Architecture}
As mentioned before, the goal of this work is to develop a driving policy by approximating through RL the action-values $Q(s,a)$ for every possible $(s,a)\in \mathcal S \times \mathcal A$ pair. Towards this direction a fully connected feed forward neural network is utilized, due to its universal function approximation property \cite{csaji2001approximation}.

Specifically, the action-values $Q(s,a)$ for each pair $(s,a)\in \mathcal S \times \mathcal A$ are approximated by using a neural network that maps a specific state $s \in \mathcal S$ to the action-values $Q(s,a_{s,i})$, where $\{a_{s,i}\}_i$ is a non empty set that contains all actions that can be selected by the policy when the agent is at state $s$. In this work, the DDQN approach is followed which utilizes two identical neural networks with two hidden layers, consist of 256 and 128 neurons respectively. The first neural network is responsible for setting the targets, while the second one is responsible for following them. The synchronization between the two neural networks is realized every 1000 epochs. For more information regarding the DDQN model please refer to \cite{van2016deep}.

\subsection{Training Set Generation and Policy Training} 
For generating $(s_t,a_t,r_t,s_{t+1})$ tuples that will be used for training the DDQN, two different microscopic traffic flow simulators are used. The first one is a custom made simulator that moves the manual driving vehicles with constant speed using the kinematics equations. The second simulator is the established SUMO\footnote{www.sumo.dlr.de/} microscopic traffic flow simulator. By exploiting traffic flow simulators driving scenarios can be simulated. For each one of the simulation steps during a simulated scenario, following the approach described in Section \ref{sec:driving_policy}, one $(s_t,a_t,r_t,s_{t+1})$ tuple can be generated using information coming directly from the simulator. 

After the collection of a set of $(s_t,a_t,r_t,s_{t+1})$ tuples the training of the RL policy is starting following the procedures described in Section \ref{sec:RL-based_driving_policy}. It should be mentioned that during policy training (and testing) we implemented a rule-based action masking \cite{liu2018elements} for changing lanes. Our choice is justified by the fact that in some driving situations, undesirable lane changes can be straightforward identified, e.g. lane changes that result to immediate collisions. In such cases undesirable lane changes are filtered out, instead of letting the agent to learn to avoid that actions. The benefits from action masking is twofold. First, it restricts the actions space, and, thus, it speeds up the learning process. Second, selection of inferior actions caused by the variance in observation will be avoided resulting to a policy that is less prone to false positives and easier to debug. Besides the aforementioned action masking, during training, no other safety mechanisms are applied on the behavior of the autonomous vehicle. On the contrary, regarding the manual driving cars, all safety mechanisms are enabled. Therefore, in case of a collision we are sure that the vehicle that caused the collision is the autonomous one. 

These are the general rules applied during the driving scenarios generation (for training and testing the RL-based driving policy) using both of the aforementioned microscopic traffic-flow simulators. Depending on the specific characteristics of each experiment extra rules may be applied. These are described in the corresponding subsections of Section \ref{sec:4}.

\subsection{Implementation details} 
For training the network we set the discount factor $\gamma=0.995$ [see relation (\ref{eq:1})], we used a memory of 2000 samples capacity, a mini-batch of 64 samples and the ADAM optimizer with learning rate $0.003$, $\beta_1 =0.9$ and $\beta_2=0.999$. The exploration factor $\epsilon_t$ at each step is annealed by
\begin{equation}
\epsilon_k = 0.01 + 0.99 e^{-\lambda k},
\end{equation}
where $k$ stands for the index of the latest training step and $\lambda$ was set equal to $7.5 \cdot 10^{-6}$. Finally, the training process started with $\epsilon_1=1.0$ and terminated when $\epsilon_k=0.01$.

\section{Safety Rules}
\label{sec:rules}
As mentioned before, no learning based driving policy can guarantee a collision free trajectory. There will always be corner cases (very rare events) that the learning algorithm will not encounter during its training phase. Therefore, it cannot be assured that the decisions corresponding to such event will be correct [for a formal proof of this result see \cite{shalev2017formal} Lemma 2]. Moreover, a vehicle might be involved in an accident without being responsible for it. For these reasons the authors of \cite{shalev2017formal} derive ad-hoc rules to guarantee responsibility-sensitive safety, that is, to guarantee that an autonomous vehicle will never cause an accident, even if it will be involved in one.

The derivation of safety rules in this work is motivated by the responsibility-sensitive framework. There is, however, a main difference between the setting in \cite{shalev2017formal} and our setting. The authors in \cite{shalev2017formal} assume that the road is occupied only by autonomous vehicles whose behaviour can be programmed. In our case, there is no such assumption. On the contrary, mixed driving scenarios are considered, where the road is occupied both by autonomous and manual driving vehicles. This implies that the behaviour of manual driving vehicles cannot be affected neither programmed. By restricting attention on vehicles that move on a highway the aforementioned assumption can be removed. This allows to assume that extreme events, such as vehicles that stop suddenly, will not occur.

Restricting attention on highways, permits also the simplification of the responsibility-safety framework by considering two types of collisions. An autonomous vehicle can cause an accident, firstly, if it moves faster than its leader and violates a minimum time gap, and, secondly, during lane changes. In the following we derive rules for avoiding these two types of collisions. Please note, that the information that can be used to derive such rules is only the information available to the autonomous vehicle, that is the positions and the velocities of the vehicles surrounding it. 

In order to avoid the first type of collisions, the minimum safety time gap $\rho_s$ that must be maintained between the autonomous vehicles and its leader needs to be estimated. Obviously, the minimum safety time gap makes sense only when the autonomous vehicle is moving faster that its leader. Let us denote as $v_{e,t}$ and as $v_{l,t}$ the longitudinal speeds of the autonomous vehicle and its leader vehicle, respectively. Also, let us denote as $d_{max}$ the maximum feasible deceleration of the autonomous vehicle. In order to avoid the first type of collisions after a time interval $\rho$, the following inequality should hold:
\begin{equation}
\label{eq:timegap}
v_{l,t} \rho - v_{e,t} \rho + \frac{1}{2}d_{max} \rho^2 > 0.
\end{equation}
Solving for $\rho$ the minimum safety time gap $\rho_s$ can be obtained by
\begin{equation}
\label{eq:min_timegap}
\rho_s = \inf \Bigg \{\rho: \rho>\frac{2(v_{e,t} - v_{l,t})}{d_{max}}\Bigg \}.
\end{equation}
Based on relation (\ref{eq:min_timegap}), the autonomous vehicle before performing an action, different than lane change actions, evaluates the minimum safety gap with respect to its leader. If the minimum time gap is violated, the autonomous vehicles decelerates with $d_{max}$ until its speed becomes equal to the speed of its leader. Otherwise, it performs the RL selected action.

Regarding the second type of collisions that can be caused by lane changes, two different cases should be considered. The autonomous vehicle should avoid collisions with its leader vehicle and with its follower vehicle in the newly selected lane. In the first case, estimates the minimum safety time gap $\rho_s$ with respect to its leader in the newly selected lane. If the minimum time gap is not violated the RL lane change action is performed. Otherwise, the autonomous vehicle selects the last action of the action set $\mathcal A$ [see Section \ref{ssec:actions}], that is to retain current lane and move with current speed, and checks for the first type of collisions. In order to avoid the collisions with its follower vehicle in the newly selected lane, the autonomous vehicle is not permitted to change lane if the follower vehicle moves faster. In this case again, the autonomous vehicle selects the last action of the action set $\mathcal A$, and checks for the first type of collisions. The rule for avoiding collisions between the autonomous vehicle and its follower is very conservative. However, since the RL-based driving policy cannot affect the behavior of the follower, and at the same time has no access to its maximum feasible deceleration (in order to relax this rule by estimating a safety time gap), such a rule is the only way to guarantee no collisions of the second type.

Although, the derived safety rules lead to a more conservative driving policy, as it can be seen in the experimental validation of the proposed approach, they permit the autonomous vehicle to advance with its desired speed and at the same time avoid collisions.

\section{Experiments}
\label{sec:4}
In this work three different sets of experiments were conducted. In the first set of experiments a simplified microscopic traffic flow simulator is utilized in order to compare the behavior of the RL-based driving policy against an optimal policy derived via Dynamic Programming. In the second set of experiments the established microscopic traffic simulator SUMO is used. Three different types of experiments are conducted. First, the behavior of the autonomous vehicle is evaluated when it is controlled by the derived RL-based policy and when it is controlled by SUMO. Second, the robustness of the derived policy with respect to measurement errors is evaluated. Finally, in the third set of experiments, the effect of vehicles that move following the RL-based policy on traffic flow is investigated. In the following the details of the experimental setup and the obtained results are presented.   

\subsection{RL-based driving policy and Dynamic Programming}
Dynamic Programming techniques can produce optimal policies assuming that no disturbances occur in the system. Due to this fact, for this set of experiments, a simplified custom made microscopic traffic simulator was developed and utilized. This simulator moves the manual driving vehicles with constant longitudinal velocity using the kinematics equations. Moreover, the manual driving vehicles are not allowed to change lanes. Despite its simplifying setting, this set of experiments allow the comparison of the RL driving policy against an optimal policy derived via Dynamic Programming. At this point it should be mentioned that for this set of experiments the ad-hoc safety rules derived in Section \ref{sec:rules} are disabled, in order to gain insights regarding the safety aspects of the RL-based driving policy.

For training the DDQN, driving scenarios of 60 seconds length were generated. In these scenarios one vehicle enters the road every two seconds, while the tenth vehicle that enters the road is the autonomous one. All vehicles enter the road at a random lane, and their initial longitudinal velocity is randomly selected from a uniform distribution ranging from $12m/s$ to $17m/s$. Finally, the desired speed of the autonomous vehicle is set equal to $21m/s$.

The RL driving policy is compared against an optimal policy derived via Dynamic Programming under four different road density values. For each one of the different densities 100 scenarios of 60 seconds length were simulated. In these scenarios, the simulator moves the manual driving vehicles, while the autonomous vehicle moves by following the RL policy and by solving a Dynamic Programming problem with 60 seconds horizon (which utilizes the same objective functions and actions as the RL algorithm). Finally, statistics regarding the number of collisions and lane changes, and the percentage of time that the autonomous vehicle moves with its desired speed for both the RL and Dynamic Programming policies are extracted. At this point it has to be mentioned that Dynamic Programming is not able to produce the solution in real time, and it is just used for benchmarking and comparison purposes. On the contrary the RL policy, at a given state can select an action very fast, since this selection corresponds to one evaluation of the neural network function at the corresponding state. 

\begin{table}[t]
	\centering
	\caption{Driving behavior evaluation of the RL and DP driving policies, in terms of total number of collision and lane changes for 100 scenarios and percentage of time that the vehicle moves with its desired speed. }
	\newcolumntype{L}[1]{>{\hsize=#1\hsize\raggedright\arraybackslash}X}%
	\newcolumntype{C}[1]{>{\hsize=#1\hsize\centering\arraybackslash}X}%
	\label{table:1}
	
	\begin{tabularx}{0.6\linewidth}{L{5.5}C{4.0}C{5.5}C{7.5}}
		\hline \hline 
		\textbf{1 veh./8 sec.} & Collisions & Lane changes & Desired speed (\%) \\ \hline
		DP policy   & 0          & 84           & 85 \\ \hline
		RL policy   & 0          & 81           & 73    \\ 
		\hline
		\hline
		\textbf{1 veh./4 sec.} &  &   &  \\ \hline
		DP policy   & 0          & 127          & 83 \\ \hline
		RL policy   & 0          & 115          & 64    \\ 
		\hline
		\hline
		\textbf{1 veh./2 sec.} &  & &  \\ \hline
		DP policy   & 0          & 120          & 87 \\ \hline
		RL policy   & 0          & 108          & 62    \\ 
		\hline
		\hline
		\textbf{1 veh./1 sec.} &  &  &  \\ \hline
		DP policy   & 0          & 70           & 72 \\ \hline
		RL policy   & 2          & 62           & 56    \\ \hline \hline

	\end{tabularx}
\end{table}

Table \ref{table:1} summarizes the results of this comparison. The four different densities are determined by the rate at which the vehicles enter the road, that is, 1 vehicle enters the road every 8, 4, 2, and 1 seconds. The RL policy is able to generate collision free trajectories, when the density is less than or equal to the density used to train the network. For larger densities, however, the RL policy produced 2 collisions every 100 scenarios. In terms of efficiency, the optimal Dynamic Programming policy is able to perform more lane changes and advance the vehicle faster.

\subsection{RL-based driving policy and SUMO policy}
In this set of experiments the behavior of the autonomous vehicle when it follows the RL policy and when it is controlled by SUMO is evaluated. The training of the RL policy took place using scenarios generated by the SUMO simulator. During the generation of scenarios, all SUMO safety mechanisms are enabled for the manual driving vehicles and disabled for the autonomous vehicle. Furthermore, the manual driving cars is not permitted to implement cooperative and strategic lane changes. Such a configuration for the lane changing behavior, impels the autonomous vehicle to implement maneuvers in order to achieve its objectives. Moreover, in order to simulate realistic scenarios two different types of manual driving vehicles are used; vehicles that want to advance faster than the autonomous vehicle and vehicles that want to advance slower. Finally, the density was equal to 600 veh/lane/hour. For the evaluation of the trained RL policy, different driving scenarios, described in the following subsections, were simulated.

\subsubsection{Evaluation of the derived RL driving policy and safety rules}
In this set of experiments different driving scenarios were simulated; i) 100 driving scenarios during which the autonomous vehicle follows the RL driving policy without the ad-hoc safety rules derived in Section \ref{sec:rules}, ii) 100 driving scenarios during which the autonomous vehicle follows the RL driving policy with the ad-hoc safety rules, iii) 100 driving scenarios during which the default configuration of SUMO was used to move forward the autonomous vehicle (cooperative and strategic lane changes are enabled for the autonomous vehicle), and iv) 100 scenarios during which the behavior of the autonomous vehicle is the same as the manual driving vehicles, i.e. it does not perform strategic and cooperative lane changes. The duration of all simulated scenarios was 60 seconds. The aforementioned scenarios' generation framework was applied for two different driving conditions. In the first one the desired speed for the slow manual driving vehicles was set to $18m/s$, while in the second one to $16m/s$. For both driving conditions the desired speed for the fast manual driving vehicles was set to $25m/s$. Furthermore, in order to investigate how the presence of uncertainties affects the behavior of the autonomous vehicle, simulated scenarios where drivers' imperfection was introduced by appropriately setting the $\sigma$ parameter in SUMO ($0\leq \sigma \leq 1$ with $\sigma=0$ to imply a perfect driver) were also used. Finally, the behavior of the autonomous vehicles was evaluated in terms of i) collision rate, and ii) average speed per scenario.

\begin{table}[t]
	\centering
	\caption{Driving behavior evaluation. \textit{SUMO default} corresponds to the default SUMO configuration, while \textit{SUMO manual} to the case where the behavior of the autonomous vehicle is the same as the  manual driving vehicles.}
	\newcolumntype{L}[1]{>{\hsize=#1\hsize\raggedright\arraybackslash}X}%
	\newcolumntype{C}[1]{>{\hsize=#1\hsize\centering\arraybackslash}X}%
	\label{table:2}
	
	\begin{tabularx}{0.6\linewidth}{L{13.5}C{4.5}C{5.5}}
		\hline \hline 
		\multicolumn{3}{ c }{\textbf{Desired speed for slow vehicles 18m/s}}\\ \hline
		& Collisions & Avg speed \\ \hline
		RL policy with rules ($\sigma=0.0$) & 0\%  & 20.62 \\ \hline
		
		RL policy w/o rules ($\sigma=0.0$) & 2\%  & 20.71 \\ \hline
		
		SUMO default ($\sigma=0.0$)    & 0\%  & 20.22    \\ \hline
		
		SUMO manual ($\sigma=0.0$)     & 0\%  & 19.48    \\ \hline
		
		RL policy with rules ($\sigma=0.5$) & 0\% & 20.08 \\ \hline
		
		RL policy w/o rules ($\sigma=0.5$) & 3\% & 20.09 \\ \hline
		
		SUMO default ($\sigma=0.5$)    & 0\% & 19.57    \\ \hline
		
		SUMO manual ($\sigma=0.5$)     & 0\%  & 19.05    \\ \hline \hline 
		\multicolumn{3}{ c }{\textbf{Desired speed for slow vehicles 16m/s}}\\ \hline
		& Collisions & Avg speed \\ \hline
		RL policy with rules ($\sigma=0.0$) & 0\%  & 19.87 \\ \hline
		
		RL policy w/o rules ($\sigma=0.0$) & 2\% & 20.04 \\ \hline
		
		SUMO default ($\sigma=0.0$)    & 0\% & 18.41    \\ \hline
		
		SUMO manual ($\sigma=0.0$)     & 0\%  & 17.47    \\ \hline
		
		RL policy with rules ($\sigma=0.5$) & 0\% & 19.81 \\ \hline
		
		RL policy w/o rules ($\sigma=0.5$) & 4\% & 19.87 \\ \hline
		
		SUMO default ($\sigma=0.5$)    & 0\% & 17.67    \\ \hline
		
		SUMO manual ($\sigma=0.5$)     & 0\%  & 17.26    \\ \hline \hline \\

	\end{tabularx}
\end{table}

Table \ref{table:2} summarizes the results of this comparison when the ad-hoc safety rules are disabled. In this way the safety levels of the RL-based driving policy can be experimentally quantified. In Table \ref{table:2}, \textit{SUMO default} corresponds to the default SUMO configuration for moving forward the autonomous vehicle, while \textit{SUMO manual} to the case where the behavior of the autonomous vehicle is the same as the manual driving vehicles. Irrespective of whether a perfect ($\sigma=0$) or an imperfect ($\sigma=0.5$) driver is considered for the manual driving vehicles, the RL policy is able to move forward the autonomous vehicle faster than the SUMO simulator, especially when slow vehicles are much slower than the autonomous one. However, it results to a collision rate of 2\%-4\%, which is its main drawback. No guarantees for collision-free trajectory is the price paid for deriving a learning based approach capable of generalizing to unknown driving situations and inferring driving actions with minimal computational cost.

\begin{figure*}[t]
	\begin{minipage}{1.0\linewidth}
		\centering
		\includegraphics[width=0.31\linewidth]{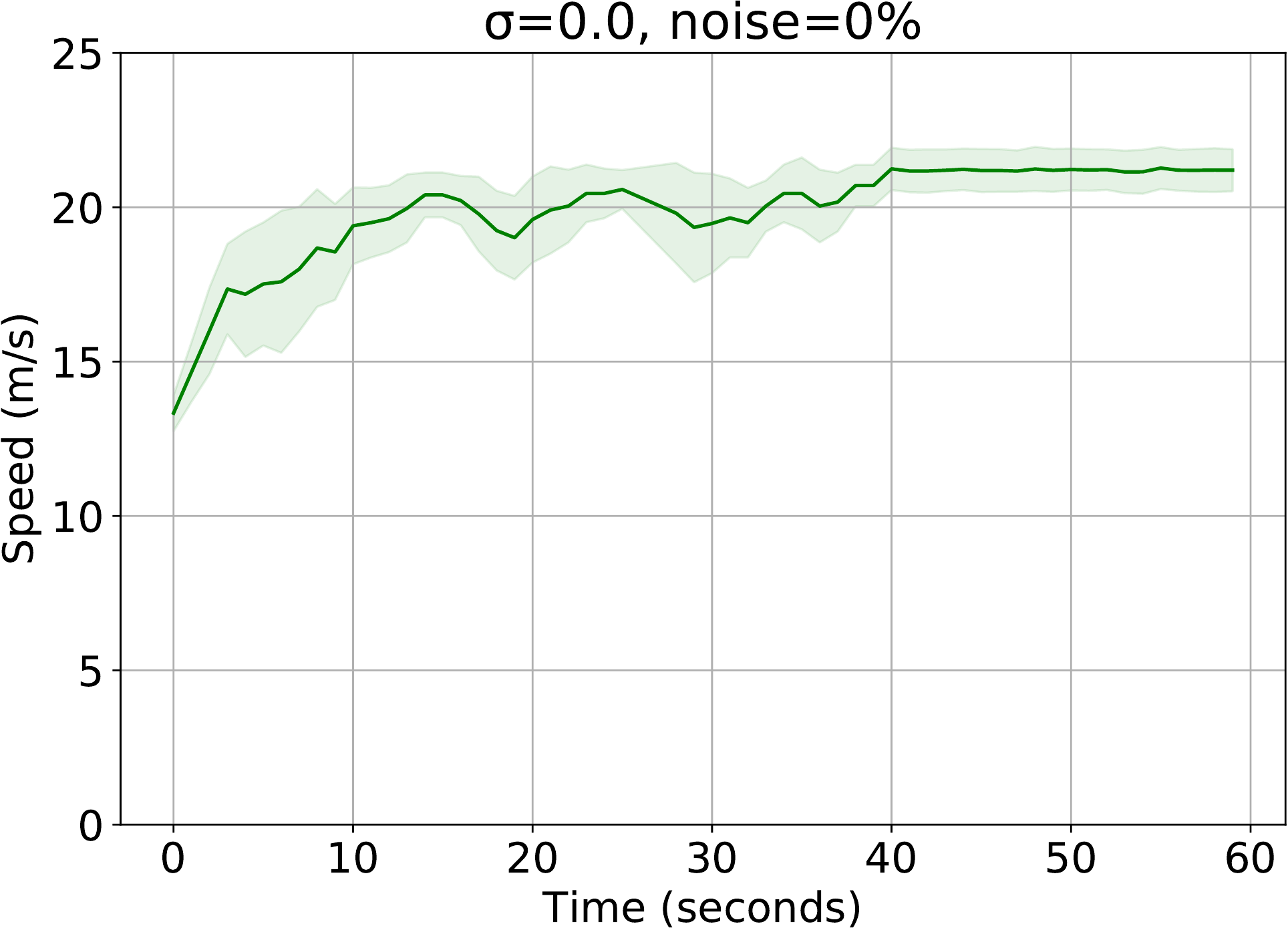} \hspace{0.15in}
		\includegraphics[width=0.31\linewidth]{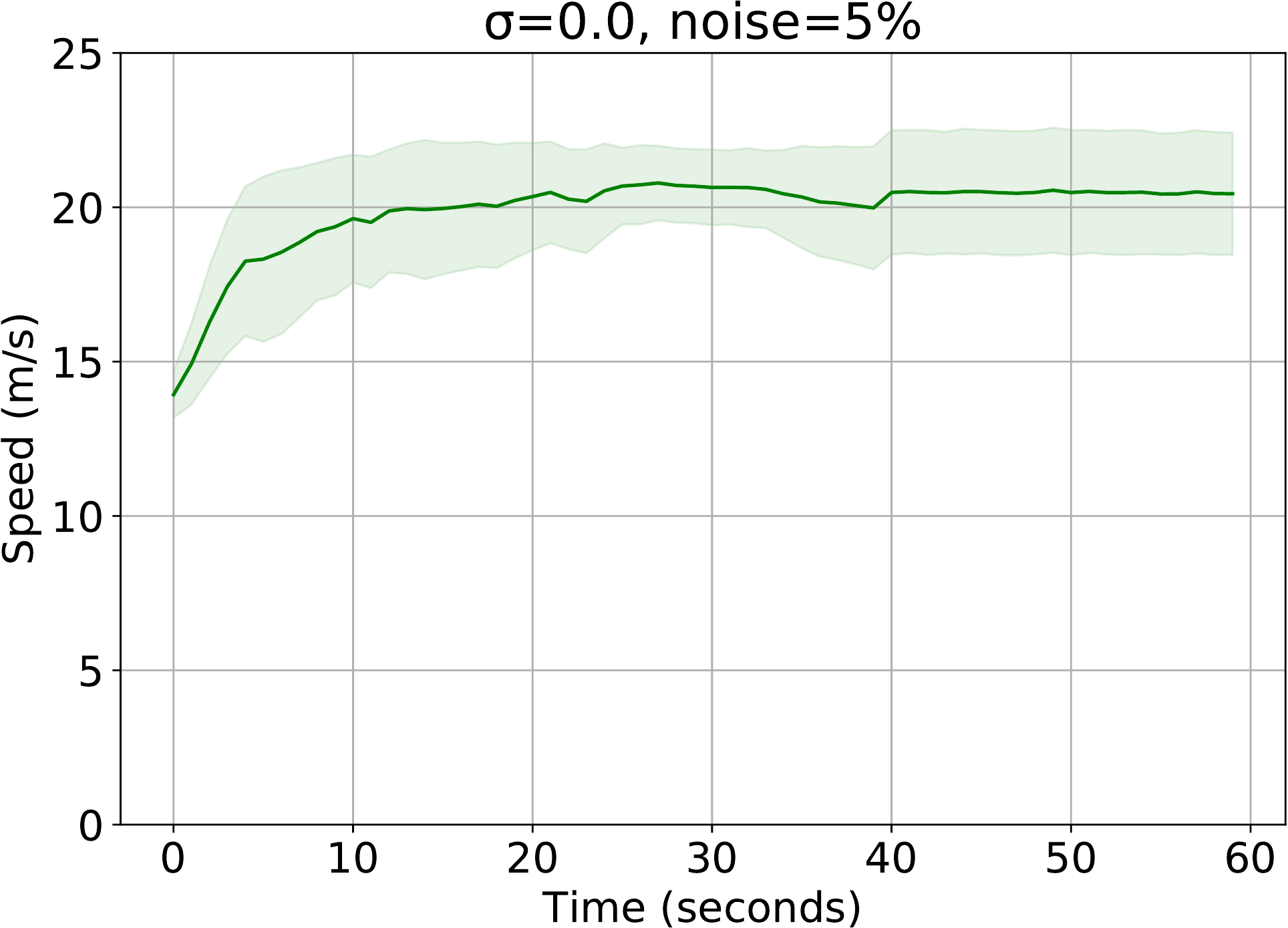} \hspace{0.15in}
		\includegraphics[width=0.31\linewidth]{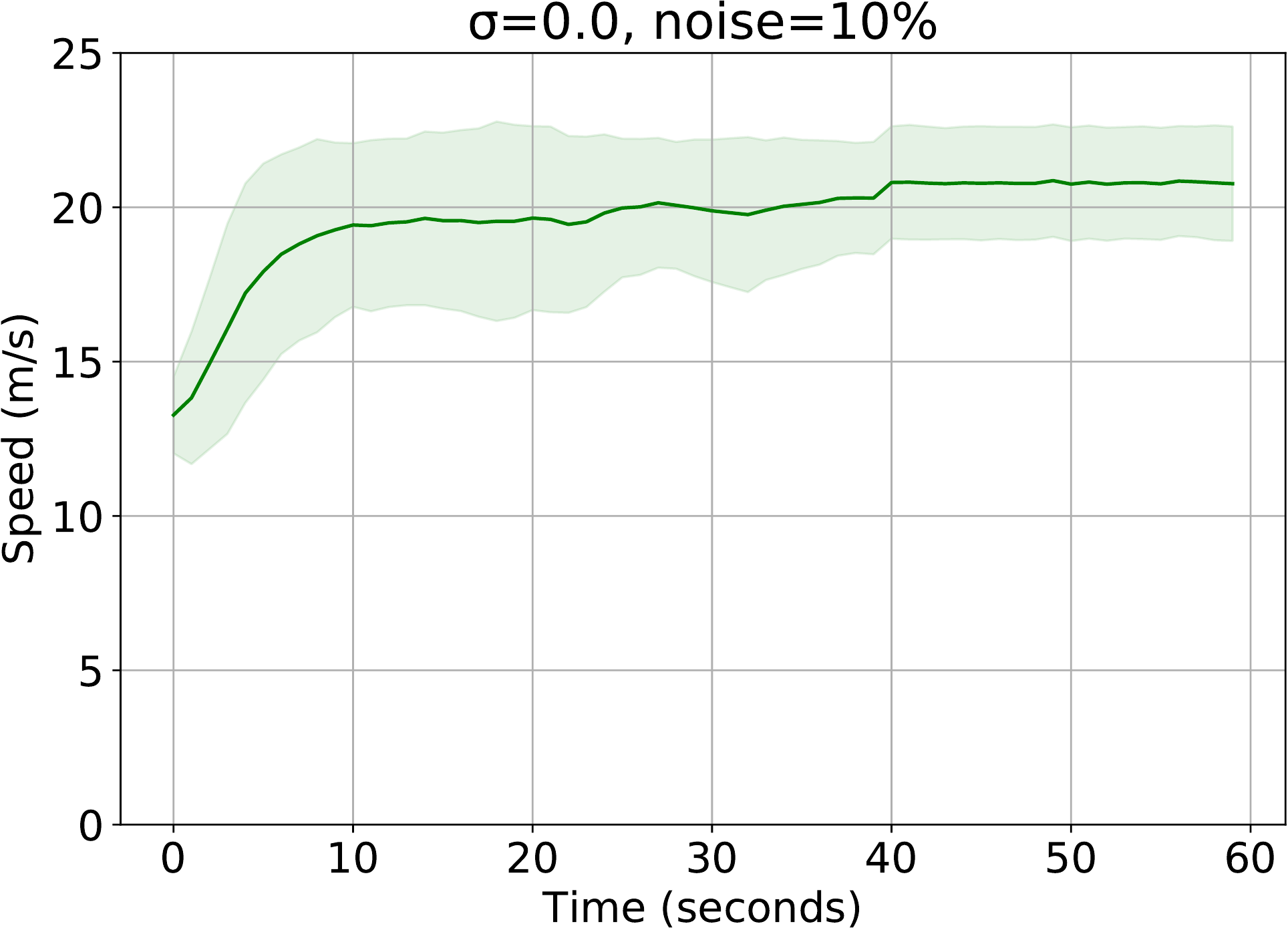}
		\vspace{0.2in}
	\end{minipage} 
	\begin{minipage}{1.0\linewidth}
		\centering
		\includegraphics[width=0.31\linewidth]{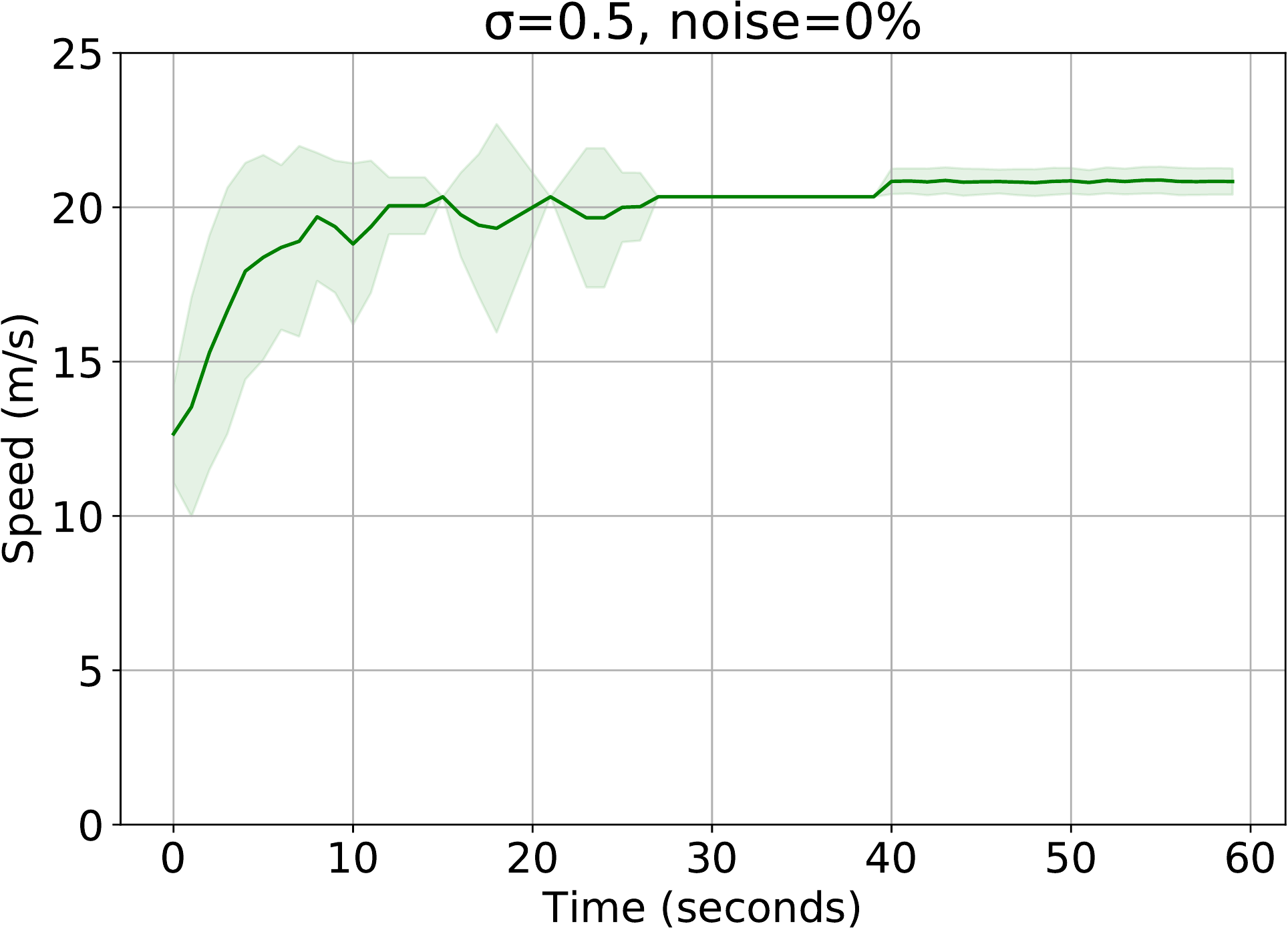} \hspace{0.15in}
		\includegraphics[width=0.31\linewidth]{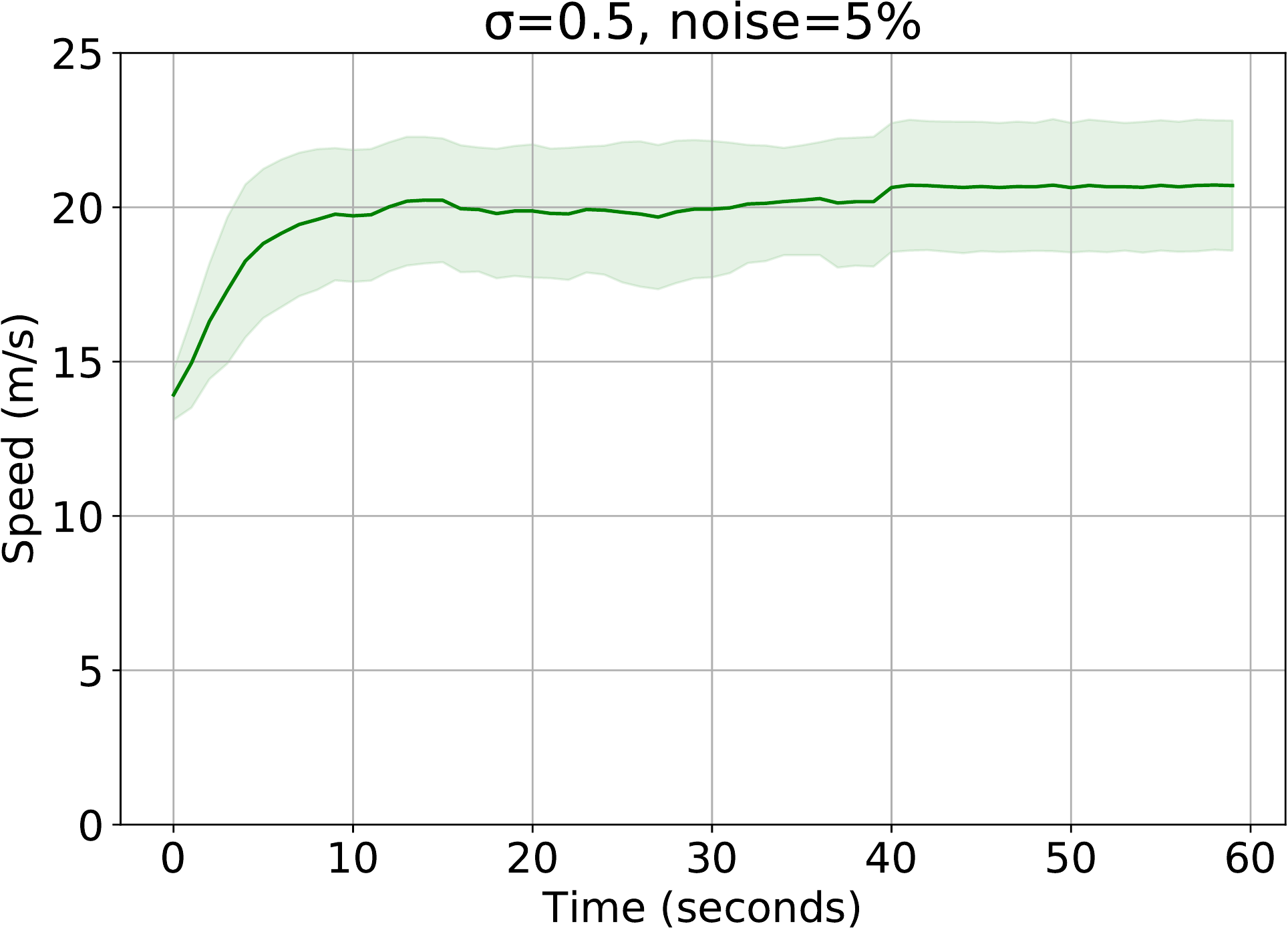} \hspace{0.15in}
		\includegraphics[width=0.31\linewidth]{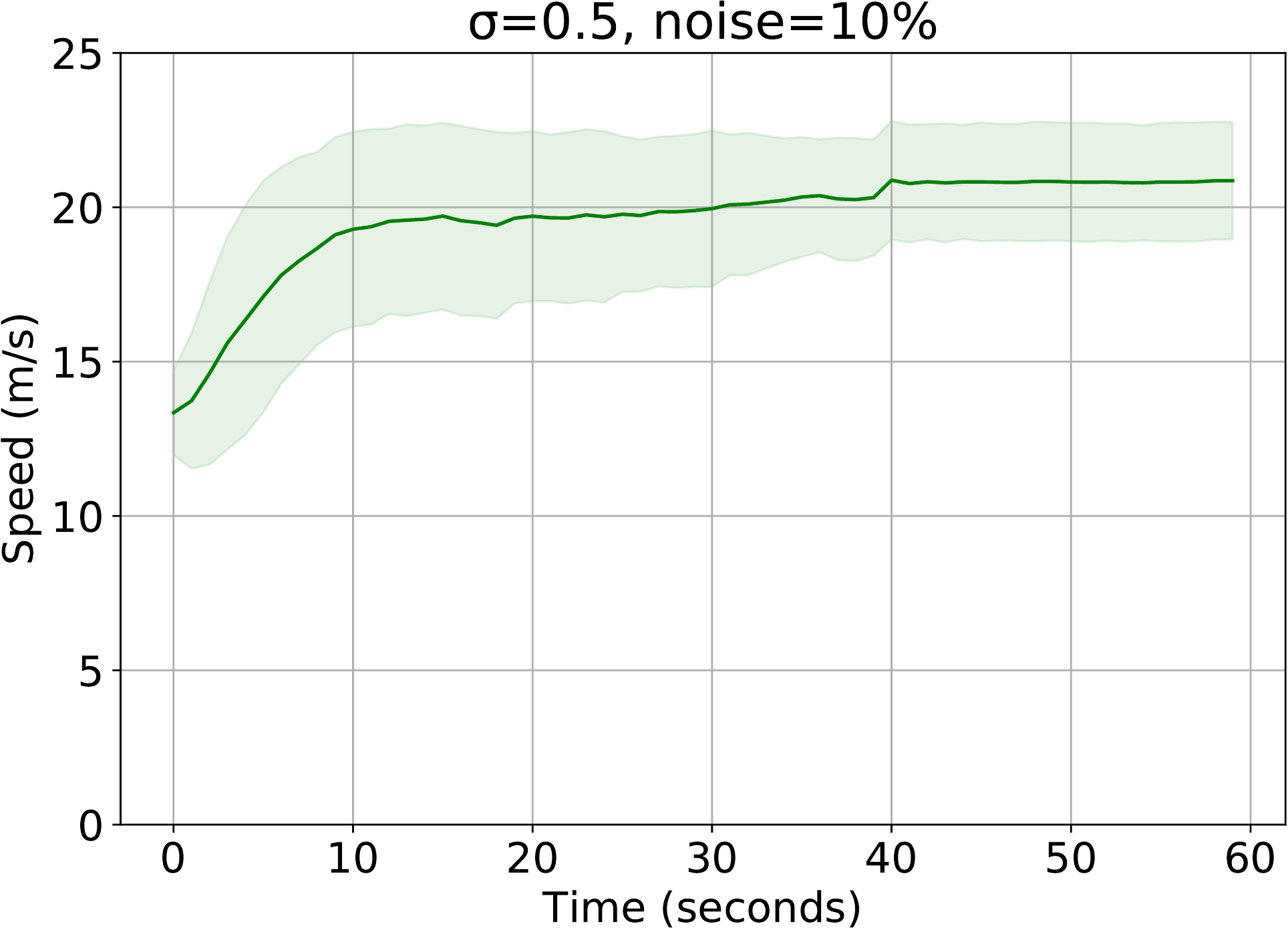}
	\end{minipage} 
	\caption{Speed trajectories for different measurements errors and driver imperfection. The solid green line represents the mean speed of the vehicle over all 100 scenarios, while the shaded area represents 1 standard deviation of the speeds below and above their mean value.}
	\label{fig:speed_trajectory}
\end{figure*}

However, when the ad-hoc safety rules are enabled the derived RL driving policy achieves to provide collision free trajectories. The average speed of the autonomous vehicle slightly decreases after the application of ad-hoc rules, but again the derived policy advances the autonomous vehicle faster than the SUMO policies. Specifically, when the speed of the slow vehicles is $18m/s$ the RL-based policy with the ad-hoc safety rules advances the autonomous vehicle 2\% and 2.6\% faster than the SUMO default policy for $\sigma=0.0$ and $\sigma=0.5$ respectively. For the case where the speed of the slow vehicles is $16m/s$, the improvement, in terms of speed, of the RL-based policy over the SUMO default policy is more significant. In particular, the RL-based policy advances the autonomous vehicle 8\% and 12\% faster than the SUMO default policy for  $\sigma=0.0$ and $\sigma=0.5$ respectively.

The aforementioned results suggest that the RL-based driving policy is not significantly more efficient than the SUMO default policy when the average speed of the manual driving vehicles is close to the desired speed of the autonomous vehicle. However, when the deviation between the desired speed of the autonomous vehicle and the average speed of the manual driving vehicles increases, the RL-based driving policy is able to advance the autonomous vehicle much faster.

\subsubsection{Evaluation of the derived RL driving policy under measurement errors}
In this set of experiments the robustness of the RL-based driving policy, with the application of ad-hoc safety rules, is evaluated with respect to measurement errors regarding the position of the manual driving vehicles. At each time step, measurement errors proportional to the distance between the autonomous vehicle and the manual driving vehicles are introduced. Two different error magnitudes were used; $\pm5\%$ and $\pm10\%$. The RL policy was evaluated in terms of collisions and average speed in 100 driving scenarios of 60 seconds length for each error magnitude. In these scenarios the desired speed of the slow vehicles is $16m/s$. Finally, for these experiments, perfect and imperfect drivers were also considered.

\begin{table}[t]
	\centering
	\caption{Driving behavior evaluation with ad-hoc safety rules when different magnitudes of measurements errors are introduced.}
	\newcolumntype{L}[1]{>{\hsize=#1\hsize\raggedright\arraybackslash}X}%
	\newcolumntype{C}[1]{>{\hsize=#1\hsize\centering\arraybackslash}X}%
	\label{table:3}
	
	\begin{tabularx}{0.6\linewidth}{L{13.5}C{4.5}C{5.5}}
		\hline \hline 
		\multicolumn{3}{ c }{\textbf{5\% Noise}}\\ \hline
		& Collisions & Avg speed \\ \hline
		RL policy with rules ($\sigma=0.0$) & 0\%  & 19.88 \\ \hline
		
		RL policy with rules ($\sigma=0.5$) & 0\%  & 19.84 \\ \hline \hline 
		\multicolumn{3}{ c }{\textbf{10\% Noise}}\\ \hline
		& Collisions & Avg speed \\ \hline
		RL policy with rules ($\sigma=0.0$) & 0\%  & 19.65 \\ \hline
		
		RL policy with rules ($\sigma=0.5$) & 0\% & 19.59 \\ \hline \hline \\

	\end{tabularx}
\end{table}

The results of this evaluation are presented in Table \ref{table:3}. Despite the introduction of noise, the RL-based driving policy is able to produce collision-free trajectories, and at the same time, retain a high speed for the autonomous vehicle. In particular, the introduction of 5\% noise does not seem to affect the average speed of the vehicle. By increasing the noise to 10\% the average speed of the vehicle slightly decreases compared to the case with noiseless measurements. Fig. \ref{fig:speed_trajectory} presents the speed trajectories of the autonomous vehicle when different drivers' imperfection and different magnitude of noises are introduced. The solid green line represents the mean speed of the vehicle over all 100 scenarios, while the shaded area represents 1 standard deviation of the speeds below and above their mean value. Irrespective of the introduced uncertainties, during the first steps of the simulation the autonomous vehicle increases its speed to reach a speed close to its desired one, and then, it retains this speed. Moreover, by increasing the noise, the deviation of the speeds over the 100 scenarios increases. This, however, is a rational behavior, since increasing the uncertainty, in terms of noisy measurements, will increase the variance during the decision making process.   

\subsubsection{Evaluation of the derived RL driving policy with unknown vehicle types}
In this set of experiments the robustness of the derived RL-based driving policy is evaluated when the road is occupied by types of vehicles that were not present during the training phase. The RL-based driving policy was trained using driving scenarios where the road was occupied by \textit{passenger} manually driving vehicles that were moving faster and slower than the autonomous vehicle. In this set of scenarios the road is occupied by the previously mentioned passenger vehicles, but also by truck, buses and motorcycles. The percentage of these types of vehicles as well as their desired speed are presented in Table \ref{table:trucks}. Under this experimental setting the robustness of the derived driving policy can be evaluated when vehicles of different sizes and different desired speeds occupy the road. Towards this direction 100 driving scenarios considering perfect drivers and 100 scenarios considering drivers' imperfections were simulated. All driving scenarios were 60 seconds long. Finally, the RL-based driving policy was evaluated in terms of collisions and average speed with which the autonomous vehicle moves forward.

\begin{table}[t]
	\centering
	\caption{Vehicle types present on the road.}
	\newcolumntype{L}[1]{>{\hsize=#1\hsize\raggedright\arraybackslash}X}%
	\newcolumntype{C}[1]{>{\hsize=#1\hsize\centering\arraybackslash}X}%
	\label{table:trucks}
	
	\begin{tabularx}{0.6\linewidth}{L{11.0}C{4.5}C{9.5}}
		\hline \hline 
		& Percentage & Maximun Speed \\ \hline
		Slow passenger vehicles & 40\%  & 16 m/s \\ \hline
		Fast passenger vehicles & 40\%  & 25 m/s \\ \hline
		Trucks                  & 5\%   & 14 m/s \\ \hline
		Buses                   & 5\%   & 16 m/s \\ \hline
		Motorcycles             & 10\%  & 21 m/s \\ \hline \hline 
		
	\end{tabularx}
\end{table}

\begin{table}[t]
	\centering
	\caption{Driving behavior evaluation with ad-hoc safety rules when different types of vehicles occupy the road.}
	\newcolumntype{L}[1]{>{\hsize=#1\hsize\raggedright\arraybackslash}X}%
	\newcolumntype{C}[1]{>{\hsize=#1\hsize\centering\arraybackslash}X}%
	\label{table:trucks_ev}
	
	\begin{tabularx}{0.6\linewidth}{L{13.5}C{4.5}C{5.5}}
		\hline \hline 
		& Collisions & Avg speed \\ \hline
		RL policy with rules ($\sigma=0.0$) & 0\%  & 19.74 \\ \hline
		
		RL policy with rules ($\sigma=0.5$) & 0\%  & 19.67 \\ \hline \hline 
		
	\end{tabularx}
\end{table}

Table \ref{table:trucks_ev} presents the RL driving policy evaluation results for the aforementioned set of experiments. By comparing these results with the results in Table \ref{table:2}, it can be seen that the average speed of the autonomous vehicle is slightly decreases by 0.13 m/s and 0.14 m/s, for $\sigma=0.0$ and $\sigma=0.5$ respectively, when types of vehicles not seen during the training phase are present in the road. This decrease is mainly due to the randomness during driving scenarios generation and not due to the presence of trucks, buses and motorcycles on the road. More importantly, the RL driving policy is able to produce collision free trajectories despite the fact that the road is occupied by types of vehicles not seen during the training phase. This is justified by two facts. First, the proposed state representation utilizes encodes about the position and the velocity of manual driving vehicles present on the road. This kind of information can be obtained and encoded for any vehicle irrespective of it type. Second, the development and application of the proposed safety rules compensates for the presence of manually driving vehicles of different sizes. It should be mentioned, however, that more realistic and accurate state representations (see Section \ref{ssec:state}) can also be utilised to explicitly encode vehicles size information in state representation.

\begin{table}[t]
	\centering
	\caption{Driving behavior evaluation with ad-hoc safety rules rainy weather conditions.}
	\newcolumntype{L}[1]{>{\hsize=#1\hsize\raggedright\arraybackslash}X}%
	\newcolumntype{C}[1]{>{\hsize=#1\hsize\centering\arraybackslash}X}%
	\label{table:rain}
	
	\begin{tabularx}{0.6\linewidth}{L{13.5}C{4.5}C{5.5}}
		\hline \hline 
		& Collisions & Avg speed \\ \hline
		RL policy with rules ($\sigma=0.5$) & 0\%  & 19.48 \\ \hline \hline 
		
	\end{tabularx}
\end{table}

\subsubsection{Evaluation of the derived RL driving policy under rainy weather conditions}
In this set of experiments the RL driving policy is evaluated under rainy weather driving conditions. Rainy weather shifts the fundamental diagram to the left, which implies, on the one hand, that the vehicles move slower, and on the other, that their maximum acceleration/deceleration becomes lower. In order to simulate rainy weather driving scenarios, the desired speed of all vehicles, except the autonomous one, was decreased by 10\%, and their maximum acceleration/deceleration by 30\%. Moreover, drivers' imperfections were also included by setting $\sigma=0.5$. Finally, different types of vehicles, that is slow and fast passenger vehicles, truck, buses and motorcycles, were also present on the road. 100 driving scenarios of 60 seconds long were simulated, and the derived driving policy was evaluated in terms of number of collisions and average speed with which the autonomous vehicle moves forward. Table \ref{table:rain} presents the result of this evaluation. The RL-based driving policy is able to produce collision free trajectories and, at the same time, move forward the autonomous vehicle with a speed larger than 19 m/s. The longitudinal velocity of the autonomous vehicle is slightly lower (0.19 m/s) than the previous experiments. This, however, is mainly caused by the decrease in overall traffic flow due to weather conditions.

\subsection{The effect of the RL-based driving policy on traffic flow}
In this set of experiments preliminary results on the effect of autonomous vehicles on the overall traffic flow are presented. Four different experiments are conducted by varying the percentage of autonomous vehicles that occupy the road. In the first experiment all vehicles are manual driving, that is the percentage of autonomous vehicles is zero. For the rest three experiments percentages of 5\%, 10\% and 15\%, respectively, are used. For each experiment 100 scenarios of 120 seconds length were simulated and for each scenario the average speed of all vehicles on the road is computed, which is an indicator of the flow; the higher the average speed the higher is the flow. For all experiments and all scenarios the desired speed of manual driving vehicles is $16m/s$ and the option for cooperative and strategic maneuvers is disabled, while the desired speed for all the autonomous vehicles is $21m/s$. In this way the behavior of the manual driving vehicles can be seen as a moving bottleneck. 

\begin{table}[t]
	\centering
	\caption{Effect of autonomous vehicles on the overall traffic flow.}
	\newcolumntype{L}[1]{>{\hsize=#1\hsize\raggedright\arraybackslash}X}%
	\newcolumntype{C}[1]{>{\hsize=#1\hsize\centering\arraybackslash}X}%
	\label{table:4}
	
	\begin{tabularx}{0.6\linewidth}{L{11.0}C{4.5}C{9.5}}
		\hline \hline 
		& Avg speed & Improvement over 0\% \\ \hline
		Autonomous vehicles 0 \% & 15.32 m/s  & 0.0\% \\ \hline
		Autonomous vehicles 5 \% & 15.41 m/s  & 0.6\% \\ \hline
		Autonomous vehicles 10\% & 16.11 m/s  & 5.1\% \\ \hline
		Autonomous vehicles 20\% & 15.91 m/s  & 1.3\% \\ \hline \hline

	\end{tabularx}
\end{table}

The results of these experiments are presented in Table \ref{table:4}. For each one of the experiments the average speed is reported. As a baseline the case where the percentage of autonomous vehicles is zero is considered, and the relative improvement of the rest of the cases (5\%, 10\%, and 15\% autonomous vehicles) against this one is also reported. The average speed for the baseline is $15.32m/s$, a little bit lower than the desired speed of the manual driving vehicles. This happens because the manual driving vehicles should satisfy the safety constraints imposed by SUMO. When the percentage of autonomous vehicles increases to 5\% the average speed of the vehicles is $15.41m/s$ resulting in a very small improvement of $0.6\%$ over the baseline. In this case the autonomous vehicles move faster than the manual driving ones. Their percentage, however, is very small, and, thus, they only slightly improve the average speed compared to the baseline. Increasing more the percentage of autonomous vehicles to 10\% results to an average speed of $16.11m/s$ and $5.1\%$ improvement compared to the baseline. Increasing, however, more the percentage of autonomous vehicles to 20\% results to an improvement of $1.3\%$ over the baseline, which is smaller than the improvement of the previous case. This mainly happens due to the selfish behavior of the autonomous vehicles.

Autonomous vehicles want to move faster than the manual driving cars, and in order to achieve that they have to perform maneuvers. Keeping the density of autonomous vehicles low permits the performance of maneuvers, and thus, the faster advancement of the autonomous vehicles. Increasing, however, the density of autonomous vehicle above a threshold, makes the performance of maneuvers more difficult, since each one autonomous vehicle, present in a limited space, wants to perform its own maneuvers in a selfish way. This results in competitive behaviors among the autonomous vehicles, which has a negative effect on the overall traffic flow. 

These preliminary results show that selfish and competitive behaviors deteriorate the overall traffic flow. Deriving an RL-based driving policy trained on scenarios where the manually driven vehicles occupy a selfish behavior will not improve the overall traffic flow. Due to limited space and the large number of maneuvers performed by the manually driven vehicles the RL training algorithm will result to a very conservative policy. In our view, the only way to improve the overall traffic flow, under mixed driving scenarios, is to derive cooperative driving policies for clusters of autonomous vehicles, in order to achieve not vehicle-centric, but overall traffic flow goals. This could be done by introducing appropriate penalty terms regarding the overall traffic flow, such as minimum travel time or average traffic flow, in the reward function. Deriving, however, cooperative RL-based driving policies for clusters of autonomous vehicles is outside the scope of this work.

\section{Discussion}
The simulation results presented in Section \ref{sec:4} indicate that the derived RL-based driving policy is more efficient, for moving forward the autonomous vehicle, than the car following model used by SUMO simulator. At the same time, the derived policy can produce collision free trajectories, and it seems to be robust under measurement errors, different types of vehicles and weather conditions. Although, the current work makes the first steps towards the exploitation of deep RL techniques for autonomous vehicles' path planning, the proposed methodology is not yet ready for real-world adoption. More complicated scenarios should be generated and utilised during the training and testing phases, such as scenarios where the autonomous vehicle is approaching a crash site ahead, heavy traffics, highway merging, emergency lane switching and night driving.

Being able to identify the limitations of the current work motivates our ongoing and future work, which comprises i) training and testing an RL-based driving policy under more complicated and realistic scenarios, ii) derive more accurate state representations by exploiting vehicle-to-vehicle and vehicle-to-infrastructure communication technologies, and iii) move from a selfish driving policy to the derivation of a cooperative driving policy in order to achieve not vehicle-centric, but overall traffic flow goals.

\section{Conclusions}
\label{sec:5}
In this work, the problem of path planning for an autonomous vehicle that moves on a freeway is considered. For addressing this problem RL techniques are employed to derive a driving policy. The driving policy is implemented using a DDQN. Two different simulators to train and validate the derived driving policy are used; a custom made microscopic traffic flow simulator and the established SUMO microscopic traffic flow simulator.  

The custom made microscopic traffic flow simulator is utilized for comparing the RL-based driving policy against an optimal policy derived via Dynamic Programming. The results of this comparison indicated that, although, Dynamic Programming can advance the autonomous vehicle faster than the RL-based driving policy, it cannot produce the trajectory in real time. Moreover, Dynamic Programming requires a priori and exact knowledge of the system dynamics in a disturbance free environment to produce an optimal solution. Due to these facts, an RL-based driving policy that incorporates the ad-hoc safety rules [see Section \ref{sec:rules}] can be proved a valuable approach for emerging driving behaviors with very low computational cost, minimal or no assumptions about the environment, and the capability to generalize to driving situations that are not known a priori.

The SUMO simulator is utilized in order to train and validate the RL-based driving policy under customary and realistic traffic scenarios. Since, no learning based approach can guarantee collision-free trajectories, ad-hoc safety rules are derived motivated by the responsibility-safety framework presented in \cite{shalev2017formal}. The derived RL-based driving policy is compared against SUMO policies with and without the introduction of uncertainties. The results of this comparison indicated that the autonomous vehicle following the RL-based policy is able to achieve higher scores. 

Finally, preliminary results regarding the effect of selfish autonomous vehicles behavior on the overall traffic flow are presented. These results suggest that, although, an individual autonomous vehicle that follows a selfish policy can achieve its goals, when multiple autonomous vehicles follow a selfish policy their impact on the overall traffic flow is negative. Selfish policies lead to competitive behaviors that deteriorate the overall traffic flow. This effect is known as the \textit{user optimum} versus \textit{system optimum} trade-off.

\section{Acknowledgement}
This research is implemented through and has been financed by the Operational Program ''Human Resources Development, Education and Lifelong Learning'' and is co-financed by the European Union (European Social Fund) and Greek national funds.

\bibliographystyle{unsrt}  
%\bibliography{refs}  %%% Remove comment to use the external .bib file (using bibtex).


\begin{thebibliography}{1}
	
	\bibitem{ziegler2014trajectory}
	Julius Ziegler, Philipp Bender, Thao Dang, and Christoph Stiller.
	\newblock Trajectory planning for bertha—a local, continuous method.
	\newblock In {\em Intelligent Vehicles Symposium (IV)}, pages 450--457. IEEE,
	2014.
	
	\bibitem{cosgun2017towards}
	Akansel Cosgun, Lichao Ma, Jimmy Chiu, Jiawei Huang, Mahmut Demir,
	Alexandre~Miranda Anon, Thang Lian, Hasan Tafish, and Samir Al-Stouhi.
	\newblock Towards full automated drive in urban environments: A demonstration
	in gomentum station, california.
	\newblock In {\em Intelligent Vehicles Symposium (IV)}, pages 1811--1818. IEEE,
	2017.
	
	\bibitem{reimer2010evaluation}
	Bryan Reimer, Bruce Mehler, and Joseph~F Coughlin.
	\newblock An evaluation of driver reactions to new vehicle parking assist
	technologies developed to reduce driver stress.
	\newblock {\em Cambridge: New England University Transportation Center,
		Massachusetts Institute of Technology}, 2010.
	
	\bibitem{donges1999conceptual}
	Edmund Donges.
	\newblock A conceptual framework for active safety in road traffic.
	\newblock {\em Vehicle System Dynamics}, 32(2-3):113--128, 1999.
	
	\bibitem{menelaou2017improved}
	Charalambos Menelaou, Stelios Timotheou, Panayiotis Kolios, and Christos~G
	Panayiotou.
	\newblock Improved road usage through congestion-free route reservations.
	\newblock {\em Transportation Research Record: Journal of the Transportation
		Research Board}, (2621):71--80, 2017.
	
	\bibitem{menelaou2017controlling}
	C~Menelaou, P~Kolios, S~Timotheou, CG~Panayiotou, and MP~Polycarpou.
	\newblock Controlling road congestion via a low-complexity route reservation
	approach.
	\newblock {\em Transportation research part C: emerging technologies},
	81:118--136, 2017.
	
	\bibitem{bast2010fast}
	Hannah Bast, Erik Carlsson, Arno Eigenwillig, Robert Geisberger, Chris
	Harrelson, Veselin Raychev, and Fabien Viger.
	\newblock Fast routing in very large public transportation networks using
	transfer patterns.
	\newblock In {\em European Symposium on Algorithms}, pages 290--301. Springer,
	2010.
	
	\bibitem{figliozzi2010vehicle}
	Miguel Figliozzi.
	\newblock Vehicle routing problem for emissions minimization.
	\newblock {\em Transportation Research Record: Journal of the Transportation
		Research Board}, (2197):1--7, 2010.
	
	\bibitem{baldacci2012recent}
	Roberto Baldacci, Aristide Mingozzi, and Roberto Roberti.
	\newblock Recent exact algorithms for solving the vehicle routing problem under
	capacity and time window constraints.
	\newblock {\em European Journal of Operational Research}, 218(1):1--6, 2012.
	
	\bibitem{bast2016route}
	Hannah Bast, Daniel Delling, Andrew Goldberg, Matthias M{\"u}ller-Hannemann,
	Thomas Pajor, Peter Sanders, Dorothea Wagner, and Renato~F Werneck.
	\newblock Route planning in transportation networks.
	\newblock In {\em Algorithm engineering}, pages 19--80. Springer, 2016.
	
	\bibitem{gillespie1997vehicle}
	Thomas~D Gillespie.
	\newblock Vehicle dynamics.
	\newblock {\em Warren dale}, 1997.
	
	\bibitem{rajamani2011vehicle}
	Rajesh Rajamani.
	\newblock {\em Vehicle dynamics and control}.
	\newblock Springer Science \& Business Media, 2011.
	
	\bibitem{zhang2013dynamic}
	Sumin Zhang, Weiwen Deng, Qingrong Zhao, Hao Sun, and Bakhtiar Litkouhi.
	\newblock Dynamic trajectory planning for vehicle autonomous driving.
	\newblock In {\em Systems, Man, and Cybernetics (SMC), 2013 IEEE International
		Conference on}, pages 4161--4166. IEEE, 2013.
	
	\bibitem{brilon2005reliability}
	Werner Brilon, Justin Geistefeldt, and Matthias Regler.
	\newblock Reliability of freeway traffic flow: a stochastic concept of
	capacity.
	\newblock In {\em Proceedings of the 16th International symposium on
		transportation and traffic theory}, volume 125143. College Park Maryland,
	2005.
	
	\bibitem{yazici2014highway}
	M~Yazici, Camille Kamga, and Kaan Ozbay.
	\newblock Highway versus urban roads: Analysis of travel time and variability
	patterns based on facility type.
	\newblock {\em Transportation Research Record: Journal of the Transportation
		Research Board}, (2442):53--61, 2014.
	
	\bibitem{shalev2017formal}
	Shai Shalev-Shwartz, Shaked Shammah, and Amnon Shashua.
	\newblock On a formal model of safe and scalable self-driving cars.
	\newblock {\em arXiv preprint arXiv:1708.06374}, 2017.
	
	\bibitem{ntousakis2016optimal}
	Ioannis~A Ntousakis, Ioannis~K Nikolos, and Markos Papageorgiou.
	\newblock Optimal vehicle trajectory planning in the context of cooperative
	merging on highways.
	\newblock {\em Transportation research part C: emerging technologies},
	71:464--488, 2016.
	
	\bibitem{goerzen2010survey}
	Chad Goerzen, Zhaodan Kong, and Bernard Mettler.
	\newblock A survey of motion planning algorithms from the perspective of
	autonomous uav guidance.
	\newblock {\em Journal of Intelligent and Robotic Systems}, 57(1-4):65, 2010.
	
	\bibitem{werling2008robust}
	Moritz Werling, Tobias Gindele, Daniel Jagszent, and Lutz Groll.
	\newblock A robust algorithm for handling moving traffic in urban scenarios.
	\newblock In {\em Intelligent Vehicles Symposium (IV)}, pages 1108--1112. IEEE,
	2008.
	
	\bibitem{fletcher2008cornell}
	Luke Fletcher, Seth Teller, Edwin Olson, David Moore, Yoshiaki Kuwata, Jonathan
	How, John Leonard, Isaac Miller, Mark Campbell, Dan Huttenlocher, et~al.
	\newblock The mit--cornell collision and why it happened.
	\newblock {\em Journal of Field Robotics}, 25(10):775--807, 2008.
	
	\bibitem{wolf2008artificial}
	Michael~T Wolf and Joel~W Burdick.
	\newblock Artificial potential functions for highway driving with collision
	avoidance.
	\newblock In {\em International Conference on Robotics and Automation (ICRA)},
	pages 3731--3736. IEEE, 2008.
	
	\bibitem{wang2015driving}
	Jianqiang Wang, Jian Wu, and Yang Li.
	\newblock The driving safety field based on driver--vehicle--road interactions.
	\newblock {\em IEEE Transactions on Intelligent Transportation Systems},
	16(4):2203--2214, 2015.
	
	\bibitem{schildbach2015scenario}
	Georg Schildbach and Francesco Borrelli.
	\newblock Scenario model predictive control for lane change assistance on
	highways.
	\newblock In {\em Intelligent Vehicles Symposium (IV)}, pages 611--616. IEEE,
	2015.
	
	\bibitem{erlien2015shared}
	Stephen~M Erlien.
	\newblock {\em Shared vehicle control using safe driving envelopes for obstacle
		avoidance and stability}.
	\newblock PhD thesis, Stanford University, 2015.
	
	\bibitem{zhang2016optimal}
	Yue~J Zhang, Andreas~A Malikopoulos, and Christos~G Cassandras.
	\newblock Optimal control and coordination of connected and automated vehicles
	at urban traffic intersections.
	\newblock In {\em 2016 American Control Conference (ACC)}, pages 6227--6232.
	IEEE, 2016.
	
	\bibitem{carvalho2014stochastic}
	Ashwin Carvalho, Yiqi Gao, St{\'e}phanie Lefevre, and Francesco Borrelli.
	\newblock Stochastic predictive control of autonomous vehicles in uncertain
	environments.
	\newblock In {\em 12th International Symposium on Advanced Vehicle Control
		(AVEC)}, 2014.
	
	\bibitem{gao2014tube}
	Yiqi Gao, Andrew Gray, H~Eric Tseng, and Francesco Borrelli.
	\newblock A tube-based robust nonlinear predictive control approach to
	semiautonomous ground vehicles.
	\newblock {\em Vehicle System Dynamics}, 52(6):802--823, 2014.
	
	\bibitem{makantasis2018motorway}
	Konstantinos Makantasis and Markos Papageorgiou.
	\newblock Motorway path planning for automated road vehicles based on optimal
	control methods.
	\newblock In {\em Transportation Research Board 97th Annual Meeting}, 2018.
	
	\bibitem{gao2010predictive}
	Yiqi Gao, Theresa Lin, Francesco Borrelli, Eric Tseng, and Davor Hrovat.
	\newblock Predictive control of autonomous ground vehicles with obstacle
	avoidance on slippery roads.
	\newblock In {\em ASME 2010 dynamic systems and control conference}, pages
	265--272. American Society of Mechanical Engineers, 2010.
	
	\bibitem{gray2012semi}
	Andrew Gray, Mohammad Ali, Yiqi Gao, J~Hedrick, and Francesco Borrelli.
	\newblock Semi-autonomous vehicle control for road departure and obstacle
	avoidance.
	\newblock {\em IFAC control of transportation systems}, pages 1--6, 2012.
	
	\bibitem{werling2012automatic}
	Moritz Werling and Darren Liccardo.
	\newblock Automatic collision avoidance using model-predictive online
	optimization.
	\newblock In {\em 51st Annual Conference on Decision and Control (CDC)}, pages
	6309--6314. IEEE, 2012.
	
	\bibitem{rasekhipour2017potential}
	Yadollah Rasekhipour, Amir Khajepour, Shih-Ken Chen, and Bakhtiar Litkouhi.
	\newblock A potential field-based model predictive path-planning controller for
	autonomous road vehicles.
	\newblock {\em IEEE Transactions on Intelligent Transportation Systems},
	18(5):1255--1267, 2017.
	
	\bibitem{papageorgiou2016feasible}
	Markos Papageorgiou, Magalene Marinaki, Konstantinos Makantasis, and Typaldos
	Panagiotis.
	\newblock A feasible direction algorithm for the numerical solution of optimal
	control problems.
	\newblock {\em Dynamic Syst. Simulation Lab., Tech. Univ. Crete, Chania,
		Greece}, 2016.
	
	\bibitem{bellman1952theory}
	Richard Bellman.
	\newblock On the theory of dynamic programming.
	\newblock {\em Proceedings of the National Academy of Sciences},
	38(8):716--719, 1952.
	
	\bibitem{bojarski2017explaining}
	Mariusz Bojarski, Philip Yeres, Anna Choromanska, Krzysztof Choromanski,
	Bernhard Firner, Lawrence Jackel, and Urs Muller.
	\newblock Explaining how a deep neural network trained with end-to-end learning
	steers a car.
	\newblock {\em arXiv preprint arXiv:1704.07911}, 2017.
	
	\bibitem{chen2017brain}
	Shitao Chen, Songyi Zhang, Jinghao Shang, Badong Chen, and Nanning Zheng.
	\newblock Brain-inspired cognitive model with attention for self-driving cars.
	\newblock {\em IEEE Transactions on Cognitive and Developmental Systems}, 2017.
	
	\bibitem{chen2015deepdriving}
	Chenyi Chen, Ari Seff, Alain Kornhauser, and Jianxiong Xiao.
	\newblock Deepdriving: Learning affordance for direct perception in autonomous
	driving.
	\newblock In {\em 2015 IEEE International Conference on Computer Vision
		(ICCV)}, pages 2722--2730. IEEE, 2015.
	
	\bibitem{xu2017end}
	Huazhe Xu, Yang Gao, Fisher Yu, and Trevor Darrell.
	\newblock End-to-end learning of driving models from large-scale video
	datasets.
	\newblock In {\em Proceedings of the IEEE Conference on Computer Vision and
		Pattern Recognition}, pages 2174--2182, 2017.
	
	\bibitem{glasmachers2017limits}
	Tobias Glasmachers.
	\newblock Limits of end-to-end learning.
	\newblock {\em arXiv preprint arXiv:1704.08305}, 2017.
	
	\bibitem{mnih2015human}
	Volodymyr Mnih, Koray Kavukcuoglu, David Silver, Andrei~A Rusu, Joel Veness,
	Marc~G Bellemare, Alex Graves, Martin Riedmiller, Andreas~K Fidjeland, Georg
	Ostrovski, et~al.
	\newblock Human-level control through deep reinforcement learning.
	\newblock {\em Nature}, 518(7540):529, 2015.
	
	\bibitem{isele2017navigating}
	David Isele, Akansel Cosgun, Kaushik Subramanian, and Kikuo Fujimura.
	\newblock Navigating intersections with autonomous vehicles using deep
	reinforcement learning.
	\newblock {\em arXiv preprint arXiv:1705.01196}, 2017.
	
	\bibitem{mukadam2017tactical}
	Mustafa Mukadam, Akansel Cosgun, Alireza Nakhaei, and Kikuo Fujimura.
	\newblock Tactical decision making for lane changing with deep reinforcement
	learning.
	\newblock In {\em submitted to International Conference on Learning
		Representations (ICLR)}, 2017.
	
	\bibitem{paxton2017combining}
	Chris Paxton, Vasumathi Raman, Gregory~D Hager, and Marin Kobilarov.
	\newblock Combining neural networks and tree search for task and motion
	planning in challenging environments.
	\newblock {\em arXiv preprint arXiv:1703.07887}, 2017.
	
	\bibitem{shalev2016safe}
	Shai Shalev-Shwartz, Shaked Shammah, and Amnon Shashua.
	\newblock Safe, multi-agent, reinforcement learning for autonomous driving.
	\newblock {\em arXiv preprint arXiv:1610.03295}, 2016.
	
	\bibitem{liu2018elements}
	Jingchu Liu, Pengfei Hou, Lisen Mu, Yinan Yu, and Chang Huang.
	\newblock Elements of effective deep reinforcement learning towards tactical
	driving decision making.
	\newblock {\em arXiv preprint arXiv:1802.00332}, 2018.
	
	\bibitem{bellman1954theory}
	Richard Bellman.
	\newblock The theory of dynamic programming.
	\newblock {\em Bulletin of the American Mathematical Society}, 60(6):503--515,
	1954.
	
	\bibitem{kanagaraj2013evaluation}
	Venkatesan Kanagaraj, Gowri Asaithambi, CH~Naveen Kumar, Karthik~K Srinivasan,
	and R~Sivanandan.
	\newblock Evaluation of different vehicle following models under mixed traffic
	conditions.
	\newblock {\em Procedia-Social and Behavioral Sciences}, 104:390--401, 2013.
	
	\bibitem{van2016deep}
	Hado Van~Hasselt, Arthur Guez, and David Silver.
	\newblock Deep reinforcement learning with double q-learning.
	\newblock In {\em AAAI}, volume~2, page~5. Phoenix, AZ, 2016.
	
	\bibitem{schaul2015prioritized}
	Tom Schaul, John Quan, Ioannis Antonoglou, and David Silver.
	\newblock Prioritized experience replay.
	\newblock {\em arXiv preprint arXiv:1511.05952}, 2015.
	
	\bibitem{watkins1992q}
	Christopher~JCH Watkins and Peter Dayan.
	\newblock Q-learning.
	\newblock {\em Machine learning}, 8(3-4):279--292, 1992.
	
	\bibitem{akai2017autonomous}
	Naoki  Akai and Luis Morales Yoichi and Takuma Yamaguchi and Eijiro Takeuchi and Yuki Yoshihara and Hiroyuki Okuda and Tatsuya Suzuki and Yoshiki Ninomiya.
	\newblock Autonomous driving based on accurate localization using multilayer LiDAR and dead reckoning.
	\newblock In {\em IEEE 20th International Conference on Intelligent Transportation Systems (ITSC)},
	2017.
	
	\bibitem{csaji2001approximation}
	Bal{\'a}zs~Csan{\'a}d Cs{\'a}ji.
	\newblock Approximation with artificial neural networks.
	\newblock {\em Faculty of Sciences, Etvs Lornd University, Hungary}, 24:48,
	2001.
	
	\bibitem{wolcott2017robust}
	Ryan W. Wolcott and Ryan M. Eustice.
	\newblock Robust LIDAR localization using multiresolution Gaussian mixture maps for autonomous driving.
	\newblock {\em The International Journal of Robotics Research}, 36(3):292--319, 2017.
	
	\bibitem{de2018fusion}
	Varuna De Silva and Jamie Roche and Ahmet Kondoz.
	\newblock Fusion of LiDAR and camera sensor data for environment sensing in driverless vehicles.
	\newblock {\em arXiv preprint arXiv:1511.05952}, 2018.
	
\end{thebibliography}
%\begin{thebibliography}{1}

%\bibitem{kour2014real}
%George Kour and Raid Saabne.
%\newblock Real-time segmentation of on-line handwritten arabic script.
%\newblock In {\em Frontiers in Handwriting Recognition (ICFHR), 2014 14th
% International Conference on}, pages 417--422. IEEE, 2014.

%\bibitem{kour2014fast}
%George Kour and Raid Saabne.
%\newblock Fast classification of handwritten on-line arabic characters.
%\newblock In {\em Soft Computing and Pattern Recognition (SoCPaR), 2014 6th
%  International Conference of}, pages 312--318. IEEE, 2014.

%\bibitem{hadash2018estimate}
%Guy Hadash, Einat Kermany, Boaz Carmeli, Ofer Lavi, George Kour, and Alon
%  Jacovi.
%\newblock Estimate and replace: A novel approach to integrating deep neural
%  networks with existing applications.
%\newblock {\em arXiv preprint arXiv:1804.09028}, 2018.

%\end{thebibliography}

\end{document}